\documentclass[10pt,twocolumn,letterpaper]{article}

\usepackage{cvpr}              
\definecolor{cvprblue}{rgb}{0.21,0.49,0.74}
\usepackage[pagebackref,breaklinks,colorlinks,allcolors=cvprblue]{hyperref}

\def\paperID{15066} 
\def\confName{CVPR}
\def\confYear{2026}

\title{Multi-GRPO: Multi-Group Advantage Estimation for Text-to-Image Generation with Tree-Based Trajectories and Multiple Rewards}

\author{
Qiang Lyu$^{1}$\ \hspace{6pt}
Zicong Chen$^{2}$\ \hspace{6pt}
Chongxiao Wang$^{3}$\ \hspace{6pt}
Haolin Shi$^{3}$\ \hspace{6pt}
Shibo Gao$^{4}$\ \hspace{6pt}
Ran Piao$^{3}$\\[2pt]
Youwei Zeng$^{5}$\ \hspace{6pt}
Jianlou Si$^{3}$\ \hspace{6pt}
Fei Ding$^{3}$\ \hspace{6pt}
Jing Li$^{3}$\ \hspace{6pt}
Chun Pong Lau$^{6}$\ \hspace{6pt}
Weiqiang Wang$^{1\dag}$\\[6pt]
$^{1}$University of Chinese Academy of Sciences\ \hspace{10pt}
$^{2}$Beihang University\ \hspace{10pt}
$^{3}$Alibaba Group\\
$^{4}$Beijing Jiaotong University\ \hspace{10pt}
$^{5}$TikTok Inc.\ \hspace{10pt}
$^{6}$City University of Hong Kong
}

\begin{document}

\twocolumn[{
\maketitle
\begin{center}
    \vspace{-0.6em} \includegraphics[width=\linewidth]{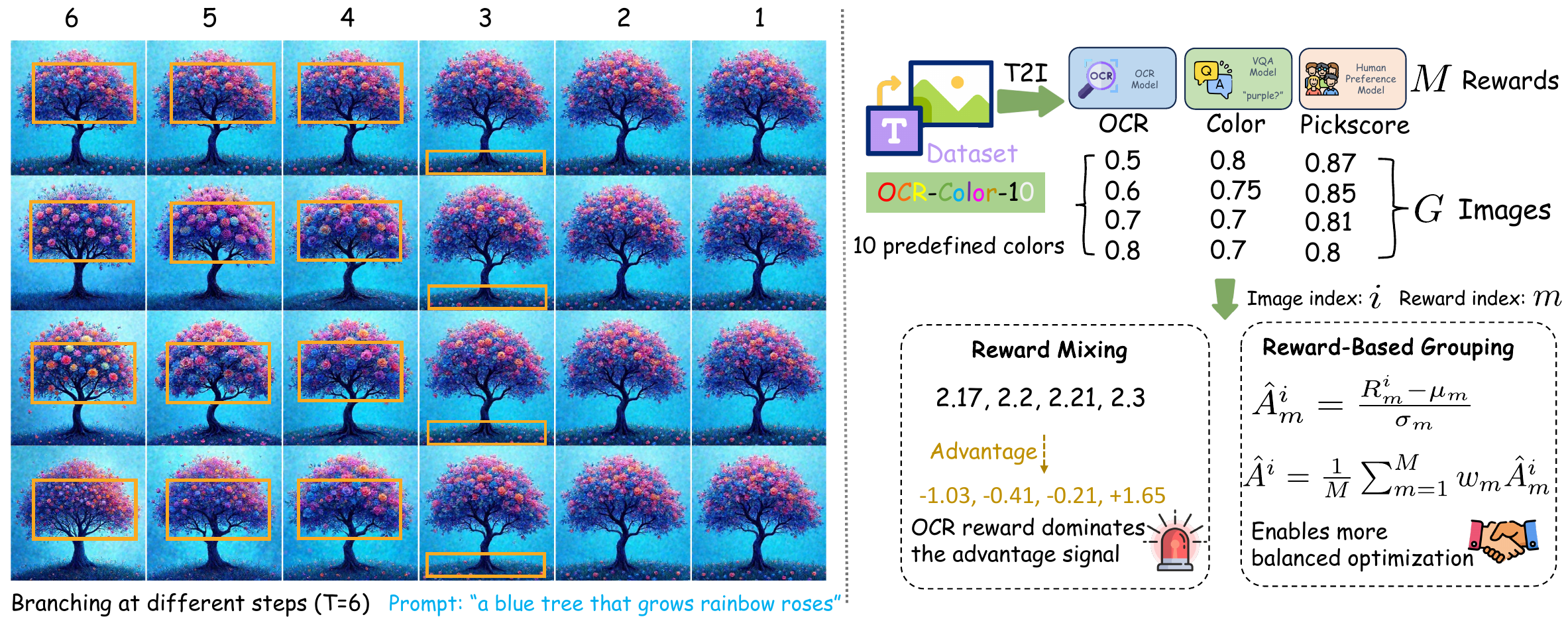}
\end{center}
\vspace{-0.9em}
\captionsetup{type=figure}
\captionof{figure}{
\textbf{(Left)} 
Early branching produces richer diversity than delayed branching, indicating that early denoising steps are more critical for exploration. Motivated by this observation, our \emph{tree-based trajectories} concentrate branching in the high-entropy early steps.
\textbf{(Right)} Illustration of the \emph{reward-mixing} problem and our \emph{reward-based grouping} solution on the curated multi-objective \textit{OCR-Color-10} dataset. 
}
\label{fig:teaser}
\vspace{0.6em}
}]

\begin{abstract}
Recently, Group Relative Policy Optimization (GRPO) has shown promising potential for aligning text-to-image (T2I) models, yet existing GRPO-based methods suffer from two critical limitations. (1) \textit{Shared credit assignment}: trajectory-level advantages derived from group-normalized sparse terminal rewards are uniformly applied across timesteps, failing to accurately estimate the potential of early denoising steps with vast exploration spaces. (2) \textit{Reward-mixing}: predefined weights for combining multi-objective rewards (e.g., text accuracy, visual quality, text color)--which have mismatched scales and variance--lead to unstable gradients and conflicting updates.
To address these issues, we propose \textbf{Multi-GRPO}, a multi-group advantage estimation framework with two orthogonal grouping mechanisms. For better credit assignment, we introduce tree-based trajectories inspired by Monte Carlo Tree Search: branching trajectories at selected early denoising steps naturally forms \emph{temporal groups}, enabling accurate advantage estimation for early steps via descendant leaves while amortizing computation through shared prefixes. For multi-objective optimization, we introduce \emph{reward-based grouping} to compute advantages for each reward function \textit{independently} before aggregation, disentangling conflicting signals.
To facilitate evaluation of multiple objective alignment, we curate \textit{OCR-Color-10}, a visual text rendering dataset with explicit color constraints. Across the single-reward \textit{PickScore-25k} and multi-objective \textit{OCR-Color-10} benchmarks, Multi-GRPO achieves superior stability and alignment performance, effectively balancing conflicting objectives. Code will be publicly available at \href{https://github.com/fikry102/Multi-GRPO}{https://github.com/fikry102/Multi-GRPO}.

\end{abstract} 
\section{Introduction}
\label{sec:intro}

Flow matching models~\cite{lipman2023flow, liu2023flow} have become a dominant paradigm for text-to-image generation~\cite{esser2024scaling, flux2024}, yet they still struggle with complex compositionality~\cite{huang2023t2i}, fine-grained attribute control, and accurate text rendering~\cite{, chen2023textdiffuser}. In parallel, online reinforcement learning, especially Group Relative Policy Optimization (GRPO)~\cite{shao2024deepseekmath}, has shown strong alignment capability in large language models (LLMs)~\cite{guo2025deepseek,yang2025qwen3} and vision language models (VLMs)~\cite{chen2025r1v}. 
Flow-GRPO~\cite{liu2025flowgrpo} first enabled stochastic exploration for flow matching via an ODE-to-SDE conversion. When applied to iterative denoising, a key challenge is shared credit assignment: a single terminal advantage is uniformly applied across all steps and obscures which decisions matter. In addition, optimizing only a single reward type can sometimes lead to reward hacking, as the model may overfit to one objective at the expense of others. At the same time, different aspects of text-to-image alignment inherently require distinct objectives, making multiple rewards necessary. However, directly combining these rewards through a weighted sum easily leads to \emph{reward mixing}, where signals with mismatched scales and variances cause one reward to dominate or drown out the others, leading to unstable or biased optimization (see Figure~\ref{fig:teaser}).

To resolve these two issues, we propose Multi-GRPO, a multi-group advantage estimation framework with two orthogonal grouping mechanisms. (1) Tree-based trajectories form temporal groups by branching at selected \textbf{early} denoising steps, giving early states Monte Carlo estimates over multiple descendants while sharing prefixes for efficiency. (2) Reward-based grouping normalizes each reward stream independently before weighted aggregation, preventing domination by high-variance objectives and yielding disentangled advantages.

In fact, some recent reinforcement learning methods for large language models have shown that incorporating tree structures can improve fine-grained credit assignment~\cite{guo2025segment,ni2025treerpo,zhang2025tree}. Our approach is conceptually aligned with these works in that we also introduce tree-based trajectories for better credit assignment. However, the text-to-image setting presents a key difference: unlike token-level decisions in LLMs, the ordering of denoising steps matters, as early steps lie in a highly uncertain and noise-dominated phase that has a disproportionately large influence on the final image.
MixGRPO~\cite{li2025mixgrpo} observes that early SDE sampling yields a more discrete and diverse latent states, highlighting the importance of exploration at the beginning of the denoising process.
Consistent with this, our preliminary experiments show that branching early shows richer diversity, whereas branching at later steps yields highly similar images, as illustrated in Figure~\ref{fig:teaser}.
This observation underlies our design choice: in text-to-image alignment, branching should be focused on the early stages where decisions play the largest role in shaping the final image.

To summarize, our contributions are three-fold: 
(i) a tree-based trajectory design that concentrates branching in early denoising steps for more accurate estimation of early-step values; 
(ii) a reward-based grouping mechanism that computes per-reward advantages prior to aggregation, preventing reward-mixing and yielding a more balanced multi-objective optimization; 
(iii) a curated visual text rendering dataset with explicit color constraints \textit{OCR-Color-10}, designed to evaluate multi-reward alignment in T2I models.
\section{Related Work}
\subsection{Diffuion and Flow Matching Models}

Generative models based on time-evolving processes, including discretized diffusion models (DDPM)\cite{ho2020denoising}, their deterministic counterpart DDIM\cite{song2020denoising}, continuous-time score-based models formulated via SDE and its corresponding ODE variant~\cite{song2021scorebased}, and flow matching models~\cite{liu2023flow,lipman2023flow}, have driven rapid advances in visual generation.
DDPM and DDIM realize the generative process through discrete stochastic or deterministic updates, whereas score-based SDE/ODE models provide continuous-time formulations whose stochastic and deterministic trajectories share the same marginal distributions. Flow matching models further advances the ODE perspective by directly learning the time-dependent velocity field, enabling high-quality synthesis with only a few deterministic ODE steps.

\subsection{Alignment for Flow Matching Models}
Reinforcement learning has recently emerged as a powerful tool for aligning text-to-image models with human preference signals. Early work such as DDPO~\cite{black2023training} demonstrated that the denoising process can be viewed as a multi-step decision-making problem, enabling effective policy-gradient optimization of diffusion models toward downstream objectives.
Building on this perspective, a new line of work adapts Group Relative Policy Optimization (GRPO)~\cite{shao2024deepseekmath} to flow matching models. Flow-GRPO~\cite{liu2025flowgrpo} and Dance-GRPO~\cite{xue2025dancegrpo} are the first to extend GRPO to flow matching by converting deterministic ODE sampling into SDEs-based sampling, enabling stochastic exploration during training.
Building on Flow-GRPO, a series of follow-up works aim to accelerate the training process, and enable precise reward assignment. MixGRPO~\cite{li2025mixgrpo} accelerates training by applying GRPO only on selected timesteps. TempFlow-GRPO~\cite{kim2025tempflow} introduces a temporally-aware GRPO framework for fine-grained credit assignment.

\subsection{Tree Structure for Reinforcement Learning}
Some recent works on reinforcement learning for large language models improve multi-step reasoning by providing more informative intermediate feedback via tree structure. SPO~\cite{guo2025segment} introduces SPO-tree for long chain-of-thought (CoT) reasoning tasks, enabling tree-based advantage estimation.
TreeRPO~\cite{ni2025treerpo} uses tree sampling to compute rewards based on step-level groups, and Tree-OPO~\cite{zhang2025tree} leverages off-policy Monte Carlo tree rollouts to guide policy optimization. Together, these approaches demonstrate that incorporating tree structure into reinforcement learning enables more effective credit assignment and leads to improved performance on complex reasoning tasks. 
Our approach is conceptually aligned with these works in that we introduce tree-based trajectories by branching at early steps for better credit assignment.
More recently, we note two closely related works, BranchGRPO~\cite{li2025branchgrpo} and DynamicTreeRPO~\cite{fu2025dynamic}, explore similar tree structures in the text-to-image domain.
Their focus, however, is on improving training efficiency by sharing prefix paths of the tree, and incorporating the sliding-window mechanism proposed from MixGRPO~\cite{li2025mixgrpo} to further accelerate training.
In contrast, our motivation for introducing a tree-based trajectories lies in obtaining more accurate estimates of the potential in the \emph{early} denoising steps.
Furthermore, our preliminary experiments indicate that branching should be performed \emph{only} at early denoising steps, as branching at later steps tends to produce nearly identical images across branches, offering limited benefit for exploring the potential of early-step decisions.

\subsection{Multiple Objective Alignment}
Recent works have explored multi-reward alignment to improve image generation. Multi-reward alignment provides richer training signals, better trade-off management among objectives, and improved alignment between prompts and generated images compared to single-reward approaches. Parrot~\cite{Parrot} formulates image generation as a multi-objective RL problem and employs Pareto-optimal selection to balance multiple rewards, including visual fidelity, textual alignment, and human preferences, thereby enhancing overall image quality. Similarly, T2I-R1~\cite{jiang2025t2i} integrates an ensemble of rewards at both semantic and token levels, guiding the model through high-level planning and fine-grained generation to produce images that are both structurally coherent and visually detailed. Distinct from prior works, we are the first to investigate multi-reward reinforcement learning in the context of Flow-GRPO.

\section{Method}
\label{sec:method}

\begin{figure*}[t]
\centering
\includegraphics[width=\linewidth]{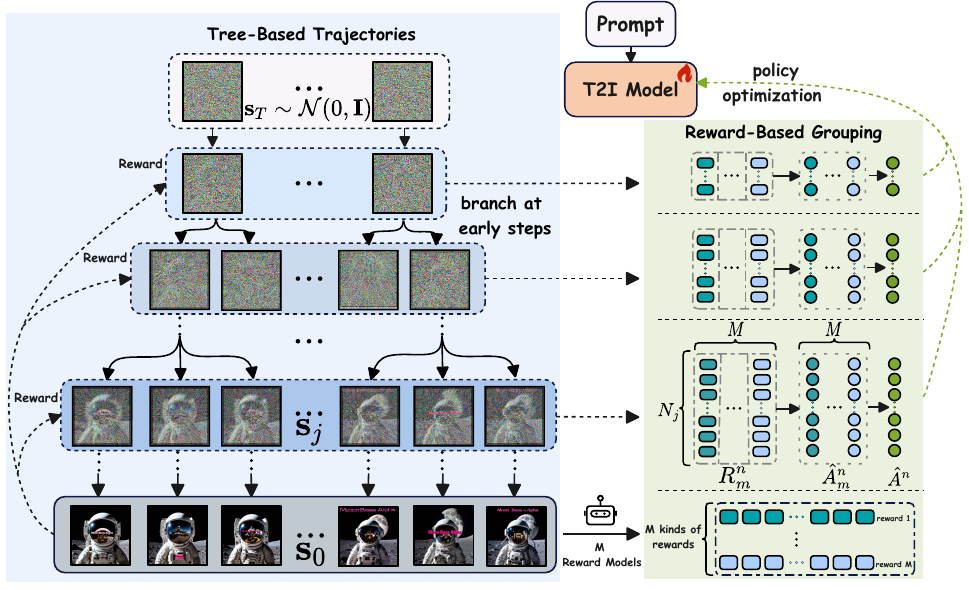}
\caption{
\textbf{Overview of Multi-GRPO.} We introduces two orthogonal grouping mechanisms to address the limitations of standard GRPO. 
\textbf{(Left)} Tree-Based Trajectories by branching at early steps: To solve the \textit{shared credit assignment} problem, we replace independent rollouts with tree-structured rollout. Early-step actions are evaluated based on a diverse set of descendant leaves, yielding more accurate estimates for critical early decisions.
\textbf{(Right)} Reward-Based Grouping: To solve the \textit{reward-mixing} problem in multi-objective optimization, we normalize advantages for each reward function independently before aggregation. This disentangles conflicting signals and prevents certain rewards from dominating the learning process. $n\in \{1,\cdots,N_j\}, m\in \{1,\cdots,M\}$. $N_j$ denotes the number of nodes at step $j$.
}\label{fig:method_overview}
\end{figure*}

\subsection{Preliminaries}
\paragraph{Flow Matching and SDE-based Sampling.}
Modern T2I models like FLUX~\cite{flux2024} and Stable Diffusion 3~\cite{esser2024scaling} are built upon the Rectified Flow~\cite{liu2023flow} framework. This framework defines the transition from a data sample $\mathbf{x}_0$ to a noise sample $\mathbf{x}_1$ via linear interpolation $\mathbf{x}_t = (1-t)\mathbf{x}_0 + t\mathbf{x}_1$, where $t \in [0, 1]$. The model trains a velocity field $\mathbf{v}_\theta(\mathbf{x}_t, t)$ to match the target velocity $\mathbf{v} = \mathbf{x}_1 - \mathbf{x}_0$. The learned velocity field defines a deterministic ordinary differential equation (ODE):
\begin{equation}
    \label{eq:ode}
    \mathrm{d}\mathbf{x}_t = \mathbf{v}_\theta(\mathbf{x}_t, t) \mathrm{d}t.
\end{equation}
However, online reinforcement learning methods like GRPO require stochastic exploration, which is absent in deterministic ODE solvers. Following~\cite{liu2025flowgrpo}, we can construct a forward stochastic differential equation (SDE) that preserves the marginal distribution $p_t(\mathbf{x}_t)$ of Eq.~\eqref{eq:ode}:
\begin{equation}
    \mathrm{d}\mathbf{x}_{t}=\left(\mathbf{v}_{\theta}(\mathbf{x}_{t}, t)+\frac{\sigma_{t}^{2}}{2}\nabla\log p_{t}(\mathbf{x}_{t})\right)\mathrm{d}t+\sigma_{t}\mathrm{~d}\mathbf{w}.
\end{equation}
Leveraging the relationship between a forward diffusion process and its time-reversal~\cite{anderson1982reverse, song2021scorebased}, the corresponding reverse SDE can be derived as:
\begin{equation}
    \label{eq:reverse_sde_with_score}
    \mathrm{d}\mathbf{x}_{t}=\left(\mathbf{v}_{\theta}(\mathbf{x}_{t}, t)-\frac{\sigma_{t}^{2}}{2}\nabla\log p_{t}(\mathbf{x}_{t})\right)\mathrm{d}t+\sigma_{t}\mathrm{~d}\bar{\mathbf{w}},
\end{equation}
where $\mathrm{d}\bar{\mathbf{w}}$ denotes the reverse Wiener process increments. In the Rectified Flow framework, the score function $\nabla\log p_{t}(\mathbf{x}_{t})$ can be derived as:
\begin{equation}
    \label{eq:score_function}
\nabla \log p_{t}(\mathbf{x}_t)=-\frac{\mathbf{x}_t}{t}-\frac{1-t}{t} \mathbf{v}_\theta(\mathbf{x}_t, t).
\end{equation}
Substituting this into Eq.~\eqref{eq:reverse_sde_with_score} yields:
\begin{align}
    \label{eq:reverse_sde}
    \mathrm{d}\mathbf{x}_t &= \left[\mathbf{v}_\theta(\mathbf{x}_t, t) + \frac{\sigma_t^2}{2t} \left(\mathbf{x}_t + (1-t) \mathbf{v}_\theta(\mathbf{x}_t, t)\right) \right] \mathrm{d}t \notag \\
    &\quad + \sigma_t \mathrm{d}\bar{\mathbf{w}}.
\end{align}
Applying the Euler-Maruyama method to discretize this SDE for a reverse step yields the final update rule: 
\begin{align}
    \label{eq:reverse_sde_step}
    \boldsymbol{x}_{t-\Delta t} &= \boldsymbol{x}_{t} - \left[\boldsymbol{v}_{\theta}(\boldsymbol{x}_{t}, t) + \frac{\sigma_{t}^{2}}{2t} \left(\boldsymbol{x}_{t} + (1-t) \boldsymbol{v}_{\theta}(\boldsymbol{x}_{t}, t)\right) \right] \Delta t \notag \\
    &\quad + \sigma_{t} \sqrt{\Delta t} \, \mathbf{z},
\end{align}
where $\mathbf{z} \sim \mathcal{N}(0, \mathbf{I})$ injects the necessary stochasticity, and $\sigma_t=a\sqrt{\frac{t}{1-t}}$ is the noise schedule with a scalar hyper-parameter $a$ controlling the noise level.

\paragraph{MDP and GRPO for T2I.}
DDPO~\cite{black2023training} proposes to formulate the iterative denoising process as a multi-step Markov Decision Process (MDP), which enables the application of reinforcement learning techniques to T2I model fine-tuning. GRPO~\cite{shao2024deepseekmath}, originally developed for language model alignment, is a lightweight, value-free policy gradient algorithm. Liu et al.~\cite{liu2025flowgrpo} adapt GRPO to flow matching models by combining it with the MDP formulation. Specifically, for a given text prompt $\mathbf{c}$, a group of $G$ trajectories $\{\tau^i\}_{i=1}^G$ are sampled from the current policy $\pi_{\theta_\text{old}}$. Each trajectory undergoes $T$ denoising steps, starting from initial noise $\mathbf{s}_T$ and progressively denoising to the final image $\mathbf{s}_0$. Each trajectory $\tau^i$ receives a terminal reward $R^i = R(\mathbf{s}_0^i, \mathbf{c})$. Actually each index $j \in \{T, T-1, \ldots, 1,0\}$ corresponds to a continuous timestep value $t_j \in [0,1]$ controlled by a timestep scheduler, with $t_T = 1$ and $t_0 = 0$. In this way, the discrete state $\mathbf{s}_j$ is connected with the continuous-time notation $\mathbf{x}_{t_j}$ in the SDE update rule (Eq.~\eqref{eq:reverse_sde_step}).

The core idea of GRPO is to estimate the advantage $\hat{A}^i$ for each trajectory by normalizing the rewards within its group:
\begin{equation}
    \label{eq:grpo_advantage}
    \hat{A}^i = \frac{R^i - \text{mean}(\{R^k\}_{k=1}^G)}{\text{std}(\{R^k\}_{k=1}^G)}.
\end{equation}
This advantage $\hat{A}^i$ is then applied uniformly to all $T$ denoising steps of trajectory $\tau^i$. The policy model is updated by maximizing the overall objective $\mathcal{J}(\theta)$:
\begin{equation}
    \label{eq:grpo_loss}
    \mathcal{J}(\theta) = \mathbb{E}_{\{\tau^i\} \sim \pi_{\theta_\text{old}}} \left[ \frac{1}{G \cdot T}\sum_{i=1}^{G} \sum_{j=1}^{T} \mathcal{L}_j^i(\theta) \right],
\end{equation}
where $\mathcal{L}_j^i(\theta)$ is the clipped, step-level objective for timestep $j$ of trajectory $\tau^i$ (transitioning from state $\mathbf{s}_j$ to $\mathbf{s}_{j-1}$):
\begin{align}
    \label{eq:grpo_step_loss}
    \mathcal{L}_j^i(\theta) = & \min\Big( r_j^i(\theta) \hat{A}^i, \operatorname{clip}(r_j^i(\theta), 1-\epsilon, 1+\epsilon) \hat{A}^i \Big) \nonumber \\
    & - \beta \mathbb{D}_{\mathrm{KL}}(\pi_{\theta}(\cdot|\mathbf{s}_j^i, \mathbf{c}) \| \pi_{\text{ref} }(\cdot|\mathbf{s}_j^i, \mathbf{c})).
\end{align}
Here, $\epsilon$ is a hyperparameter introduced in~\cite{schulman2017proximal} for stablizing training, and $r_j^i(\theta)$ is the ratio of action probabilities under the current and old policy at timestep $j$:
\begin{equation}
    \label{eq:grpo_ratio}
    r_j^i(\theta) = \frac{\pi_\theta(\mathbf{s}_{j-1}^i \mid \mathbf{s}_j^i, \mathbf{c})}{\pi_{\theta_{\text{old}}}(\mathbf{s}_{j-1}^i \mid \mathbf{s}_j^i, \mathbf{c})}.
\end{equation}

\subsection{Tree-Based Trajectories}
\label{sec:tree_trajectories}

\paragraph{Motivation.}
The standard GRPO approach~\cite{liu2025flowgrpo} relies on independent rollouts, where all timesteps within a trajectory share a single terminal signal. This shared credit assignment is insufficient for accurately estimating the potential of \emph{early steps} (e.g., $t=T-1,T-2, \dots$). These early steps operate in a high-entropy latent space (near pure noise) with vast degrees of freedom, and decisions made here have critical impacts on the final image. Thus a single trajectory-level advantage provides inadequate signal to reliably estimate the value of these critical early-step actions.

\begin{figure}[t]
    \centering
    \includegraphics[width=\linewidth]{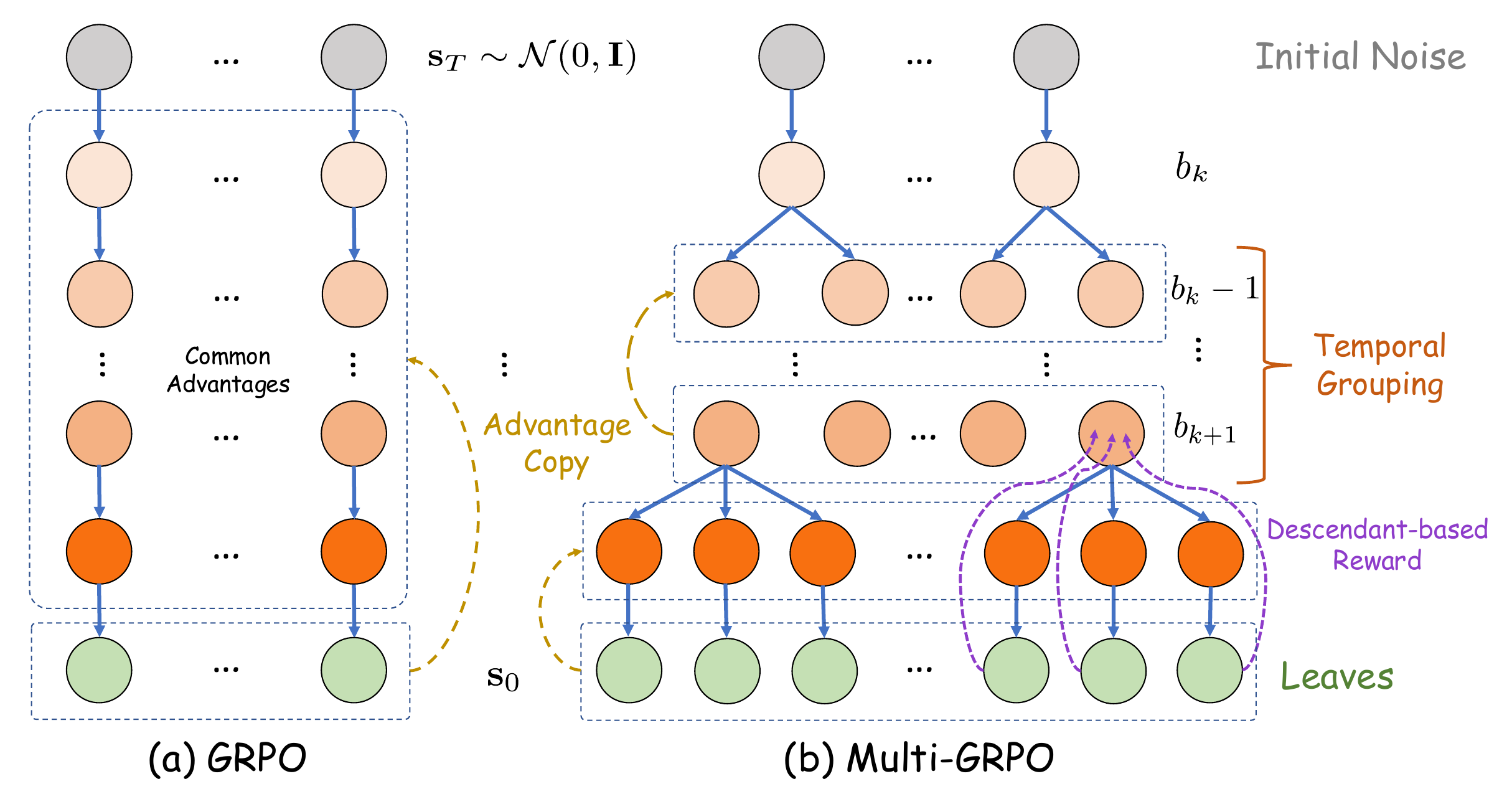}
    \caption{
        \textbf{Illustration of Tree-based Trajectories.}
        \textbf{(Left)} Standard GRPO: It uses independent rollouts. Trajectory-level advantages derived are uniformly applied across timesteps,, leading to poor credit assignment for early steps.
        \textbf{(Right)} Our method (Tree-Based Trajectories): We construct a tree by branching at selected early steps (e.g., $b_k,b_{k+1}$). This allows an early-step state to be evaluated by a diverse set of descendant leaves, providing a more reliable Monte Carlo estimate of its value.
    }
    \label{fig:tree}
\end{figure}

\paragraph{Tree-Structured Rollout.}
To better estimate the value of early-step actions, we draw inspiration from Monte Carlo Tree Search (MCTS) and replace independent rollouts with \emph{tree-structured rollout}. This process is visualized in Figure~\ref{fig:tree}. Instead of generating $G$ trajectories independently from start to finish, we construct \emph{tree-based trajectories} by branching at predefined steps, allowing multiple sub-trajectories to share their common prefixes.

Specifically, we define a \emph{branching schedule} $\mathcal{B}$, a dictionary mapping selected step indices to branching factors. Formally, $\mathcal{B} = \{b_1{:}B_1, b_2{:}B_2, \ldots, b_K{:}B_K\}$ specifies that at step $b_k \in \{T, T-1, \ldots, 1,0\}$, each existing trajectory splits into $B_k$ children, where $b_1 > b_2 > \cdots > b_K$ (branching from early to late steps). 
For notational convenience in describing \emph{temporal grouping}, we introduce two auxiliary indices: $b_0 = T$ (the initial step) and $b_{K+1} = 0$ (the final step, where no further branching occurs). For a given prompt, the tree starts with the same initial noise and branch into $B_0$ nodes at step $T$, where $B_0$ represents the degree of parallel sampling (in practice, $B_0$ is the total batch size). The total number of leaf trajectories is $N_0 = \prod_{k=0}^{K} B_k$.

To implement branching efficiently, we exploit the additive structure of the SDE update rule (Eq.~\eqref{eq:reverse_sde_step}). At each branching step $j \in \{b_1, b_2, \ldots, b_K\}$ with branching factor $\mathcal{B}[j]$ and corresponding continuous timestep $t_j$, we compute the velocity field $\mathbf{v}_\theta(\mathbf{x}_{t_j}, t_j)$ \textbf{once} through a transformer model. Then we compute the mean of the next-step distribution, which is deterministic and can be reused:
\begin{equation}
    \label{eq:drift_term}
    \boldsymbol{\mu} = \boldsymbol{x}_{t_j} - \left[\boldsymbol{v} + \frac{\sigma_{t_j}^{2}}{2t_j} \left(\boldsymbol{x}_{t_j} + (1-t_j) \boldsymbol{v}\right) \right]\Delta t_j,
\end{equation}
where $\Delta t_j = t_j - t_{j-1}$, $\boldsymbol{u}$ denotes $ \boldsymbol{u}_{\theta}(\boldsymbol{x}_{t_j}, t_j)$ and $\boldsymbol{v} $ denotes $ \boldsymbol{v}_{\theta}(\boldsymbol{x}_{t_j}, t_j)$ for brevity. We reuse this shared mean $\boldsymbol{\mu}$ across all $\mathcal{B}[u]$ child branches and sample $\mathcal{B}[u]$ independent noise vectors $\mathbf{z}^{(i)} \overset{\text{i.i.d.}}{\sim} \mathcal{N}(0, \mathbf{I})$ for $i=1, \ldots, \mathcal{B}[u]$:
\begin{equation}
    \label{eq:branching_step}
    \mathbf{x}_{t_{j-1}}^{(i)} = \boldsymbol{\mu} + \sigma_{t_j} \sqrt{\Delta t_j} \, \mathbf{z}^{(i)}.
\end{equation}
This amortizes the expensive transformer computation across all $\mathcal{B}[u]$ branches.

\paragraph{Temporal Grouping.}
This tree-based trajectoires naturally enables temporal grouping for more accurate credit assignment. At any step $j \in \{T, T-1, \ldots, 0\}$, the tree contains $N_j$ distinct states (nodes), which we denote as $\{\mathbf{s}_j^{n}\}_{n=1}^{N_j}$. Each state $\mathbf{s}_j^{n}$ corresponds to a unique path from the root and has a set of descendant leaf trajectories. Let $\mathcal{D}_j^{n} \subseteq \{1, \ldots, N_0\}$ denote the set of leaf indices that descend from state $\mathbf{s}_j^{n}$. We assign a descendant-based reward to $\mathbf{s}_j^{n}$ by averaging the terminal rewards of its descendant leaves:
\begin{equation}
    \label{eq:descendant_reward}
    R_j^{n} = \frac{1}{|\mathcal{D}_j^{n}|} \sum_{\ell \in \mathcal{D}_j^{n}} R_0^{\ell},
\end{equation}
where $R_0^{\ell} = R(\mathbf{s}_0^{\ell}, \mathbf{c})$ is the terminal reward of the $\ell$-th leaf.

We compute advantages separately for temporal segments between consecutive branching steps. For all steps $j$ in the segment $(b_{k}, b_{k+1}]$ (i.e., from step $b_{k}-1$ to step $b_{k+1}$), the number of nodes remains constant at $N_j = N_{b_{k+1}}$. We normalize the descendant-based rewards of all states at step $b_{k+1}$ to compute advantages:
\begin{equation}
    \label{eq:temporal_advantage}
    \hat{A}_{b_{k+1}}^{n} = \frac{R_{b_{k+1}}^{n} - \mu_{b_{k+1}}}{\sigma_{b_{k+1}} },
\end{equation}
where $\mu_{b_{k+1}}$ and $\sigma_{b_{k+1}}$ are the mean and standard deviation of states at step $b_{k+1}$.
This advantage $\hat{A}_{b_{k+1}}^{n}$ is then applied uniformly to all steps $j \in (b_{k}, b_{k+1}]$ along the path through state $\mathbf{s}_{b_{k+1}}^{n}$. This tree-based design improves credit assignment in two ways: (1) early-step states are evaluated via Monte Carlo estimation over multiple descendant outcomes (Eq.~\eqref{eq:descendant_reward}), providing more reliable estimates than a single trajectory; (2) states at each branching step are normalized by temporal grouping (Eq.~\eqref{eq:temporal_advantage}), so that states at different denoising stages are compared within their respective groups.

\subsection{Reward-Based Grouping}
\label{sec:reward_grouping}

\paragraph{Problem of Reward-Mixing.}
T2I alignment often requires optimizing for multiple objectives simultaneously. The naive approach is to compute a weighted sum of rewards $R^i = \sum_{m=1}^M w_m R^i_m$ for each trajectory $i$ ($M$ is the number of reward functions), then normalize this mixed reward to compute the advantage (Eq.~\eqref{eq:grpo_advantage}). However, when reward functions have mismatched scales and variances, normalizing the weighted sum causes rewards with larger magnitudes or variances to dominate the advantage signal, suppressing contributions from other objectives.

\begin{figure}[t]
    \centering
    \includegraphics[width=\linewidth]{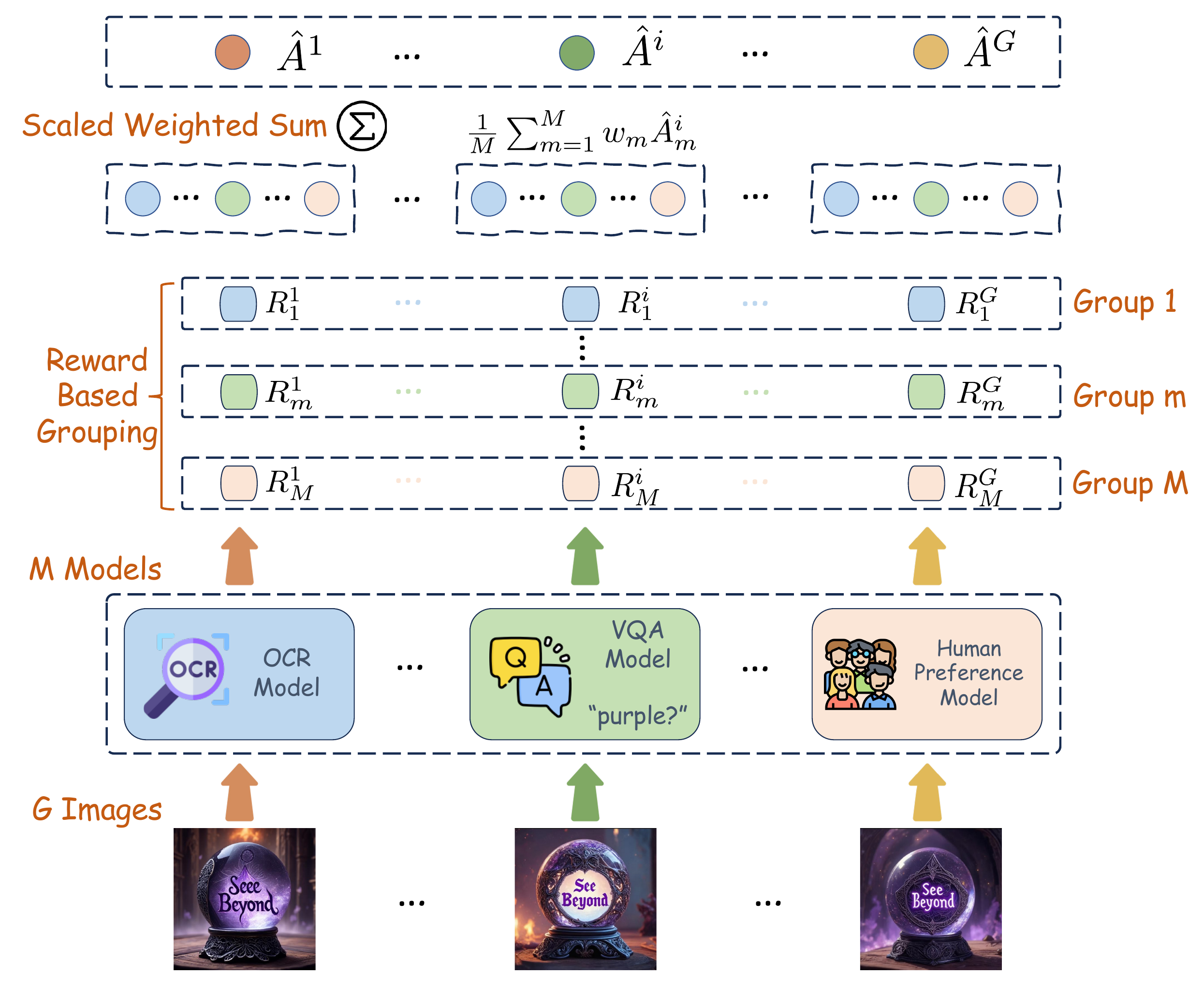}
    \caption{
        \textbf{Illustration of Reward-Based Grouping.} To address the \textit{reward-mixing} problem in multi-objective optimization, we avoid mixing rewards before normalization. Instead, each reward type (e.g., OCR,Color) is normalized independently against its own group statistics (mean and std of ``\emph{Group} $m$") to compute a disentangled advantage. The final advantage $\hat{A}^i $ is then formed by a scaled weighted sum of these individual advantages $\hat{A}^i_m$.
    }
    \label{fig:multiple_rewards}
\end{figure}

\paragraph{Reward-Based Grouping.}
\label{3.3-reward}
To address this, we propose \emph{reward-based grouping} (see Figure~\ref{fig:multiple_rewards}), which computes advantages for each reward type independently before aggregation, rather than first mixing rewards into a weighted sum. Specifically, for a group of $G$ trajectories with $M$ reward functions, each trajectory $i$ receives a reward vector $(R^i_1, \dots, R^i_M)$. We compute advantages for each reward type independently:
\begin{equation}
    \label{eq:decomposed_advantage}
    \hat{A}^i_m = \frac{R^i_m - \mu_m}{\sigma_m},
\end{equation}
where $\mu_m$ and $\sigma_m$ are the mean and standard deviation of only the $m$-th reward across the group. The final advantage is the scaled weighted sum:
\begin{equation}
    \label{eq:final_advantage}
    \hat{A}^i = \frac{1}{M}\sum_{m=1}^M w_m \hat{A}^i_m.
\end{equation}
This ensures each reward contributes to the policy gradient relative to its own scale and variance, providing a stable, disentangled learning signal for multi-objective optimization.

Although dividing the aggregated advantage by $M$ is a natural scaling choice—ensuring that the overall magnitude remains comparable as the number of rewards increases, our experiments show that a variant without this $1/M$ factor seems perform better. Since $M$ may be large in general multi-objective settings, we keep the scaled version for better scalability and discuss the unscaled variant in the supplementary material.

\section{Experiments}

\subsection{Experiment Setup}
\paragraph{Task 1 (Single-Objective Alignment on \textit{PickScore-25k}):} For the single-reward alignment task, we follow Flow-GRPO~\cite{liu2025flowgrpo} and use PickScore~\cite{kirstain2023pick} as our reward model. This experiment is conducted using FLUX.1-Dev model~\cite{flux2024}, which we train 2000 steps. It is trained on the \textit{PickScore-25k} dataset introduced in Flow-GRPO, which contains \emph{25,432} training prompts and \emph{2,048} test prompts. This task aims to align T2I models with human preferences.

\paragraph{Task 2 (Multi-Objective Alignment on \textit{OCR-Color-10}):} This task evaluates the model's ability to follow complex, multi-faceted instructions. This experiment is conducted using SD3.5-M model~\cite{esser2024scaling}, which we train 4200 steps. To create a challenging multi-objective benchmark, we curate the \textit{OCR-Color-10} dataset. Specifically, we extend the visual text rendering task from Flow-GRPO~\cite{liu2025flowgrpo, chen2023textdiffuser} by adding explicit color constraints to the generated text.
We predefined a set of 10 target colors and used Qwen3-30B-A3B-Instruct-2507~\cite{yang2025qwen3} to identify prompts in the original dataset that already contained explicit color descriptions, retaining only those that matched our specified color set. For prompts without a color reference, we randomly assigned one of the 10 colors to achieve an approximately balanced color distribution across the final dataset. The resulting dataset contains \emph{17,641} prompts for training and \emph{914} prompts for testing.

This task requires the model to optimize three potentially conflicting rewards simultaneously:
(i) $R_{\text{ocr}}$ (Text Fidelity):
Measures the textual accuracy by computing the normalized edit distance between OCR-extracted and ground-truth text.
(ii) $R_{\text{color}}$ (Text-Color Consistency):
Assesses whether the rendered text matches the target color by querying a vision-language model~\cite{wang2022gitgenerativeimagetotexttransformer} with color-specific prompts (e.g., ``purple text?") and using the normalized “Yes” probability as the score.
(iii) $R_{\text{pickscore}}$ (Image Quality):
Evaluates overall image quality and text–image alignment using PickScore~\cite{kirstain2023pick}.

\begin{table}[ht]
\caption{Results on the sinlge-objective \textit{PickScore-25k} dataset.}
\label{tab:single_reward_task}
\centering
\begin{tabular}{lc}
\toprule
Model & PickScore $\uparrow$ \\
\midrule
FLUX.1-Dev (Base Model) & 22.04 \\
Flow-GRPO (Sequential Rollout) & 23.65 \\
Multi-GRPO (\textbf{Tree-based Trajectories}) & \textbf{24.24} \\
\bottomrule
\end{tabular}
\end{table}

\subsection{Implementation Details}
At each training iteration, we set $B_0$ to 8 . Under the default branching rule $\mathcal{B}=\{T{-}2:3,\ T{-}4:2\}$, branching is performed for target
timesteps $t = T{-}2$ and $t = T{-}4$, i.e., at the 2nd and 4th denoising steps
from the end (see
Sec.~\ref{sec:supp-branch-impl} in the supplementary material). This yields a total of 48 leaf nodes per iteration for reward computation. Following Flow-GRPO~\cite{liu2025flowgrpo}, the noise-level hyperparameter $\alpha$ is set to 0.7, and all models are trained at an image resolution of 512.
We also follow Flow-GRPO in adopting model-specific denoising schedules. For \textbf{Flux.1-Dev}, we use 6 training steps and 28 inference (sampling) steps. For \textbf{SD3.5-M}, we use 10 training steps and 40 inference steps. These step counts define the total timesteps $T$. 
All experiments are conducted on 8$\times$ NVIDIA H20 GPUs.

\subsection{Main Results}

\paragraph{Task 1: Tree-Based Trajectories Improve Single-Objective Alignment.}
We begin by evaluating the impact of tree-based trajectories in a single-reward setting using the \textit{PickScore-25k} dataset. As shown in Table~\ref{tab:single_reward_task}, sequential \textbf{Flow-GRPO} already improves over the FLUX.1-Dev base model. However, when applying our \textbf{Tree-Based Trajectories}, the model achieves a substantially higher PickScore, reaching 24.24. This demonstrates that replacing sequential rollouts with tree-structured rollouts provides better preference alignment. 

\paragraph{Task 2: Multi-GRPO Achieves More Balanced Multi-Objective Optimization.}
We evaluate our full method on the \textit{OCR-Color-10} benchmark. Flow-GRPO acts as a naive-sum baseline: it first mixes the rewards into a unified reward,
$R_{\text{total}} = w_1 R_{\text{ocr}} + w_2 R_{\text{color}} + w_3 R_{\text{pick}}$,
using the default setting $w_1 = w_2 = w_3 = 1$, and then computes a single advantage. As shown in Table~\ref{tab:main_multi}, this approach yields clear improvements on both $R_{\text{ocr}}$ and $R_{\text{color}}$, but provides only a marginal gain on PickScore from 22.24 to 22.59 (note that $R_{\text{pick}}$ corresponds to PickScore normalized by 26), indicating that the PickScore objective is largely under-optimized. In contrast, our Multi-GRPO—combining Tree-Based Trajectories with Reward-Based Grouping (Eq.~\ref{eq:decomposed_advantage}, Eq.~\ref{eq:final_advantage})—achieves consistent improvements across all three objectives. The slightly lower $R_{\text{ocr}}$ (by just 0.0046) may stem from mild conflicts among the three objectives, yet overall our method achieves a more balanced optimization across all metrics.

\begin{table}[ht]
\caption{Results on the multi-objective \textit{OCR-Color-10} dataset.}
\label{tab:main_multi}
\centering
\begin{tabular}{lcccc}
\toprule
Model & $R_{\text{color}}$ $\uparrow$ & $R_{\text{ocr}}$ $\uparrow$ & $R_{\text{pick}}$ $\uparrow$ & PickScore $\uparrow$ \\
\midrule
SD3.5-M  & 0.8141 & 0.6542 & 0.8554 & 22.24 \\
Flow-GRPO & 0.9471 & \textbf{0.9428} & 0.8689 & 22.59 \\
\textbf{Multi-GRPO} & \textbf{0.9516} & 0.9382 & \textbf{0.8966} & \textbf{23.31} \\
\bottomrule
\end{tabular}
\end{table}
\begin{figure}[h]
\centering
\includegraphics[width=\linewidth]{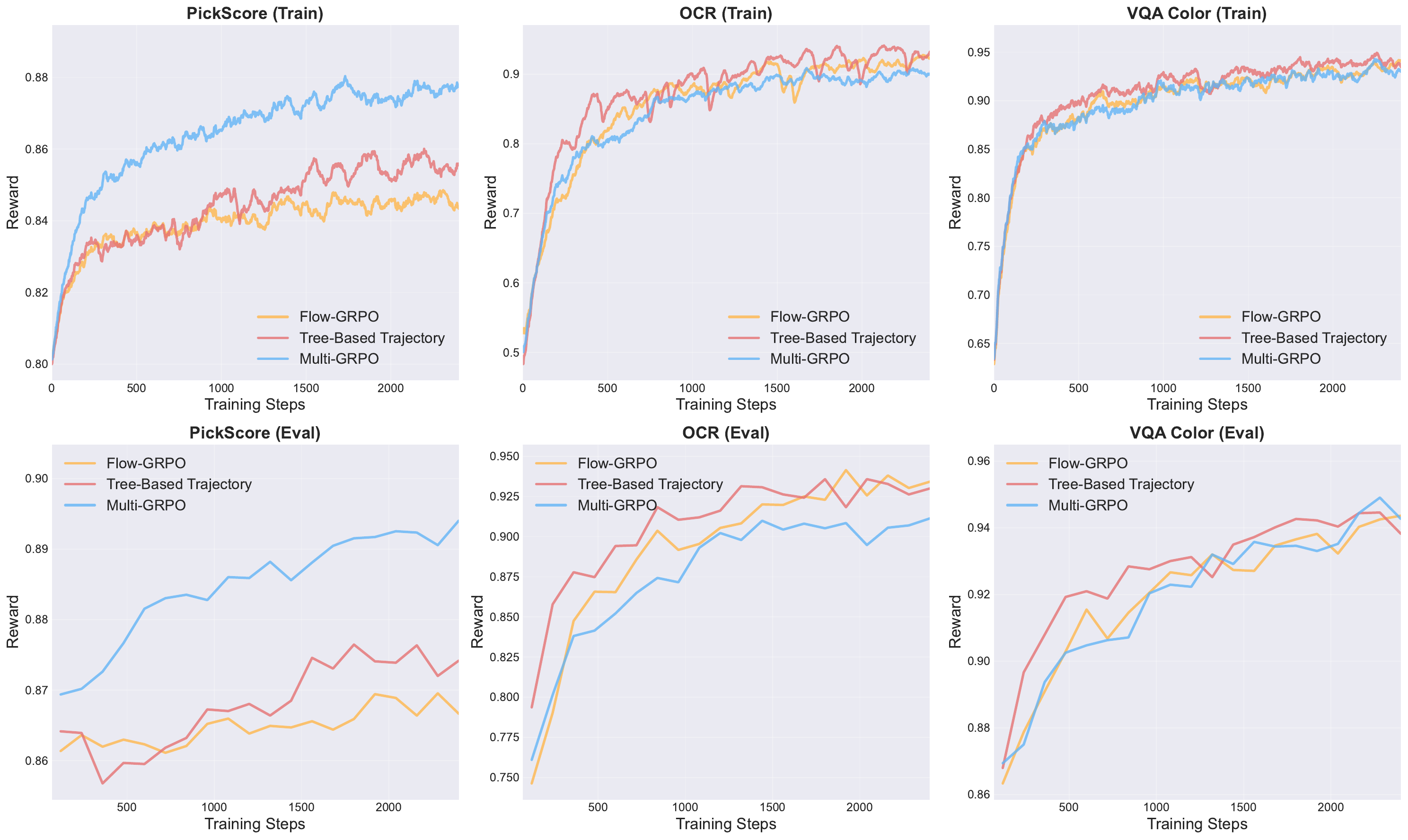}
\caption{
Ablation on the \textit{OCR-Color-10} dataset. 
Relative to the Flow-GRPO baseline, Tree-Based Trajectories show a generally upward trend across all three rewards, although the gain in $R_{\text{pick}}$ remains modest. 
Adding Reward-Based Grouping on top of Tree-Based Trajectories further boosts PickScore and leads to a more balanced improvement overall, forming the full Multi-GRPO approach.
}
\label{fig:ablation_figure}
\end{figure}
\subsection{Ablation Analysis}

Figure~\ref{fig:ablation_figure} shows the effect of adding each component on the \textit{OCR-Color-10} dataset. 
Starting from the Flow-GRPO baseline, introducing \textbf{Tree-Based Trajectories} generally improves all three rewards, indicating that tree-structured rollouts are beneficial even under \emph{multi-objective settings}, although the improvement on $R_{\text{pick}}$ is still limited. 
Adding \textbf{Reward-Based Grouping} on top of Tree-Based Trajectories further boosts PickScore and yields a more balanced improvement across the three rewards. 
Combining both components (Multi-GRPO) results in the most balanced overall performance.

\paragraph{Impact of Branching Rule.}
We study how the branching schedule affects performance. Our default rule applies early splits, $\mathcal{B}=\{T{-}2:3,\ T{-}4:2\}$, and we compare it to a late-branching variant with the same total number of leaf nodes, $\mathcal{B}=\{T{-}3:2,\ T{-}7:3\}$. As shown in Table~\ref{tab:ablation_branch}, early branching yields better performance across all three rewards. This supports that branching at early denoising steps produces more informative trajectories and improves optimization.

\begin{table}[h]
\centering
\caption{Comparison of early vs.\ late branching rules. Both configurations produce the same number of leaf nodes, but early branching yields better performance across all reward dimensions.}
\label{tab:ablation_branch}
\begin{tabular}{lccc}
\toprule
Branching Rule & $R_{\text{color}}$ $\uparrow$ & $R_{\text{ocr}}$ $\uparrow$ & $R_{\text{pick}}$ $\uparrow$ \\
\midrule
$\{T{-}2{:}3,\ T{-}4{:}2\}$  & \textbf{0.9516} & \textbf{0.9382} & \textbf{0.8966} \\
$\{T{-}3{:}2,\ T{-}7{:}3\}$  & 0.9449 & 0.9033 & 0.8912 \\
\bottomrule
\end{tabular}
\end{table}

\begin{figure}[t]
    \centering
    \includegraphics[width=\linewidth]{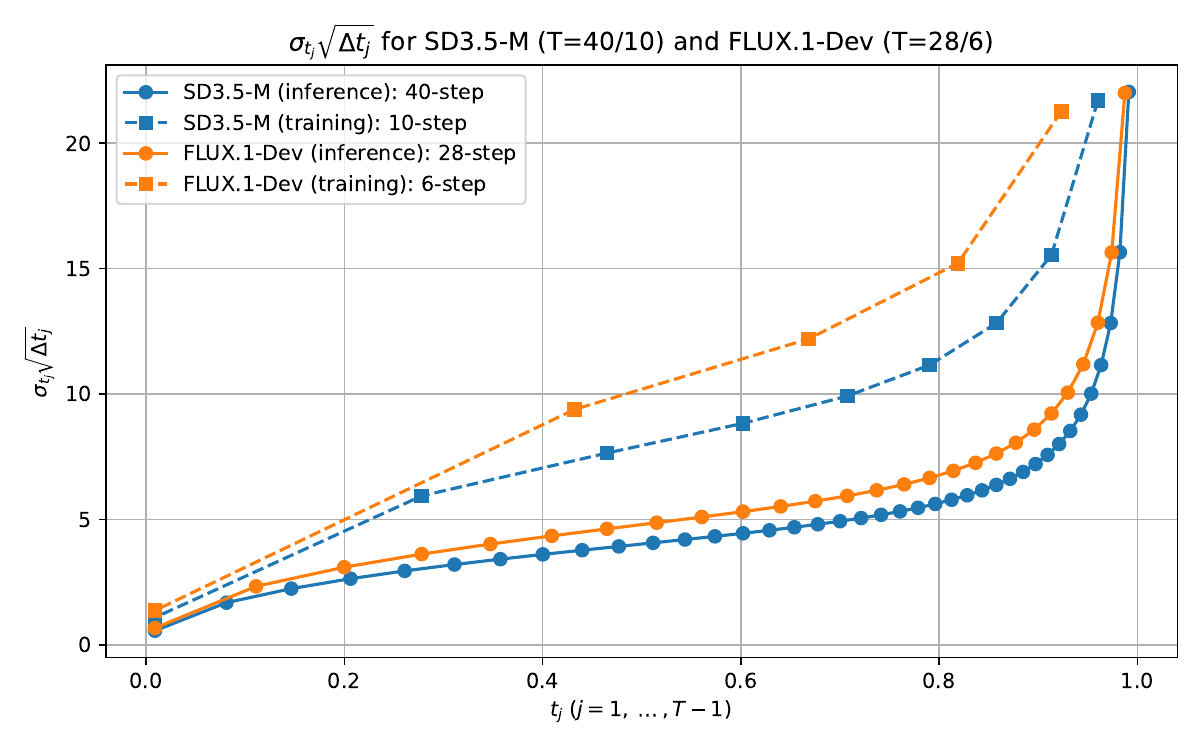}
    \caption{Noise strength ($\sigma_{t_j}\sqrt{\Delta t_j}$) across different timesteps for multiple denoising schedules (T=6, 10, 28, 40). Early timesteps exhibit higher noise, where more stochastic exploration is needed for effective credit assignment.}
    \label{fig:noise_strength}
\end{figure}
\subsection{Additional Analysis}
\paragraph{Noise Strength across Timesteps.}
As defined in the SDE update rule (Eq.~\ref{eq:reverse_sde_step}), the noise coefficient $\sigma_{t} \sqrt{\Delta t}$ controls the strength of the injected Gaussian noise $\mathbf{z}$. To better understand the stochasticity introduced during denoising, we visualize this coefficient across timesteps in Figure~\ref{fig:noise_strength}. This coefficient results in higher noise levels early in the denoising process (when $t$ is close to 1), reflecting the high-entropy state near pure noise. In contrast, noise diminishes as $t$ approaches 0, aligning with the more deterministic nature of the later steps. This visualization supports our core argument for concentrating branching in early stages, where stochastic exploration is most critical for effective credit assignment.

\paragraph{Necessity for Multi-Objective Optimization.}
Optimizing a single reward can introduce a significant risk of \emph{reward hacking}, where the model over-optimizes one metric at the expense of other, often conflicting, objectives such as overall visual quality.
\begin{figure}[t]
\centering
\includegraphics[width=\linewidth]{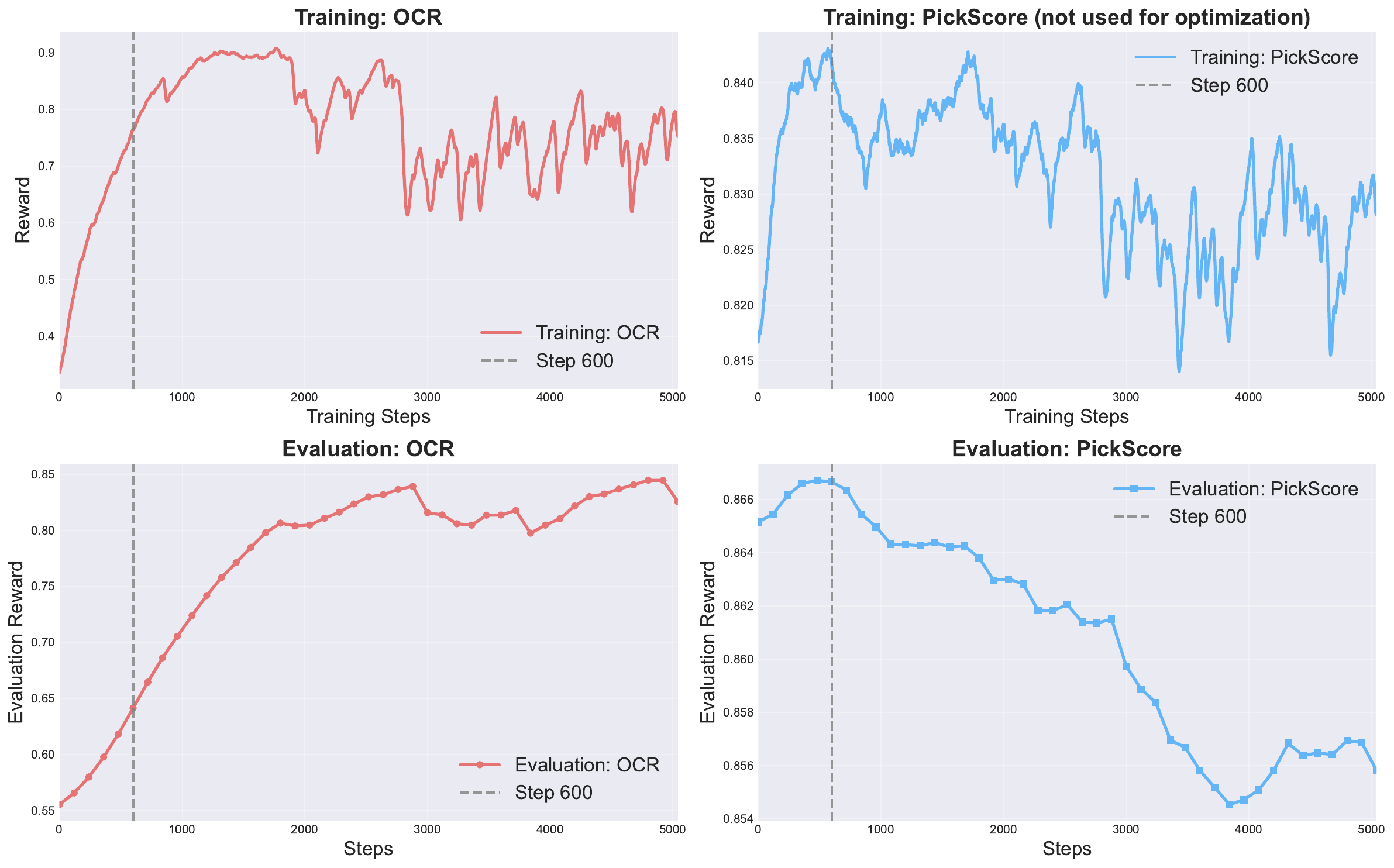}
\caption{Optimizing only for the $R_{\text{ocr}}$ (text fidelity) reward leads to a significant drop in the PickScore (aesthetic quality), illustrating possible reward hacking in single-objective settings and highlighting the need for balanced, multi-objective optimization.}
\label{fig:single_objective_hacking}
\end{figure}
As shown in Figure~\ref{fig:single_objective_hacking}, we investigate this effect by training a model to maximize $R_{\text{ocr}}$ (text fidelity) exclusively, using the original OCR dataset from Flow-GRPO~\cite{liu2025flowgrpo}. 
While text accuracy improves, the overall perceptual quality, as measured by PickScore, rapidly degrades. This collapse underscores the necessity of  multi-objective optimization.

\section{Conclusion}
In this work, we propose \textbf{Multi-GRPO}, a multi-group advantage estimation framework that addresses two key limitations of GRPO-based text-to-image alignment: shared credit assignment and reward mixing. 
\emph{Tree-based trajectories} provide better credit assignment by evaluating early denoising steps through multiple descendant leaves. 
\emph{Reward-based grouping} computes advantages for each objective independently, avoiding reward mixing and enabling more balanced multi-objective optimization. 
To facilitate evaluation of multi-objective alignment, we curate the \textit{OCR-Color-10} dataset. 
Experiments show that Multi-GRPO achieves stronger performance in single-objective settings and more balanced optimization in multi-objective settings.

\clearpage
{
    \small
    \bibliographystyle{ieeenat_fullname}
    \bibliography{main}

@String(ECCV= {Eur. Conf. Comput. Vis.})

@String(ICLR = {Int. Conf. Learn. Represent.})

@String(ECCV  = {ECCV})

@String(ICLR  = {ICLR})

@article{shao2024deepseekmath,
  title={DeepSeekMath: Pushing the Limits of Mathematical Reasoning in Open Language Models},
  author={Shao, Zhihong and Wang, Peiyi and Zhu, Qihao and Xu, Runxin and Song, Junxiao and Bi, Xiao and Zhang, Haowei and Zhang, Mingchuan and Li, Y.K. and Wu, Y. and Guo, Daya},
  journal={arXiv preprint arXiv:2402.03300},
  year={2024}
}

@inproceedings{liu2023flow,
  title={Flow Straight and Fast: Learning to Generate and Transfer Data with Rectified Flow},
  author={Liu, Xingchao and Gong, Chengyue and Liu, Qiang},
  booktitle={International Conference on Learning Representations (ICLR)},
  year={2023}
}

@inproceedings{ho2020denoising,
  title={Denoising Diffusion Probabilistic Models},
  author={Ho, Jonathan and Jain, Ajay and Abbeel, Pieter},
  booktitle={Advances in Neural Information Processing Systems (NeurIPS)},
  pages={6840--6851},
  year={2020}
}

@misc{flux2024,
    author={Black Forest Labs},
    title={FLUX},
    year={2024},
    howpublished={\url{https://github.com/black-forest-labs/flux}},
}

@inproceedings{liu2025flowgrpo,
  title={Flow-GRPO: Training Flow Matching Models via Online RL},
  author={Liu, Jie and Liu, Gongye and Liang, Jiajun and Li, Yangguang and Liu, Jiaheng and Wang, Xintao and Wan, Pengfei and Zhang, Di and Ouyang, Wanli},
  booktitle={Advances in Neural Information Processing Systems (NeurIPS)},
  year={2025}
}

@article{li2025branchgrpo,
  title={BranchGRPO: Stable and Efficient GRPO with Structured Branching in Diffusion Models},
  author={Li, Yuming and Wang, Yikai and Zhu, Yuying and Zhao, Zhongyu and Lu, Ming and She, Qi and Zhang, Shanghang},
  journal={arXiv preprint arXiv:2509.06040},
  year={2025}
}

@article{xue2025dancegrpo,
  title={DanceGRPO: Unleashing GRPO on Visual Generation},
  author={Xue, Zeyue and Wu, Jie and Gao, Yu and Kong, Fangyuan and Zhu, Lingting and Chen, Mengzhao and Liu, Zhiheng and Liu, Wei and Guo, Qiushan and Huang, Weilin and Luo, Ping},
  journal={arXiv preprint arXiv:2505.07818},
  year={2025}
}

@article{li2025mixgrpo,
  title={MixGRPO: Unlocking Flow-based GRPO Efficiency with Mixed ODE-SDE},
  author={Li, Junzhe and Cui, Yutao and Huang, Tao and Ma, Yinping and Fan, Chun and Yang, Miles and Zhong, Zhao},
  journal={arXiv preprint arXiv:2507.21802},
  year={2025}
}

@inproceedings{song2021scorebased,
  title={Score-Based Generative Modeling through Stochastic Differential Equations},
  author={Song, Yang and Sohl-Dickstein, Jascha and Kingma, Diederik P. and Kumar, Abhishek and Ermon, Stefano and Poole, Ben},
  booktitle={International Conference on Learning Representations (ICLR)},
  year={2021}
}

@inproceedings{lipman2023flow,
  title={Flow Matching for Generative Modeling},
  author={Lipman, Yaron and Chen, Ricky T. Q. and Ben-Hamu, Heli and Nickel, Maximilian and Le, Matt},
  booktitle={International Conference on Learning Representations (ICLR)},
  year={2023}
}

@article{anderson1982reverse,
  title={Reverse-time diffusion equation models},
  author={Anderson, Brian DO},
  journal={Stochastic Processes and their Applications},
  volume={12},
  number={3},
  pages={313--326},
  year={1982},
  publisher={Elsevier}
}

@inproceedings{esser2024scaling,
  title={Scaling Rectified Flow Transformers for High-Resolution Image Synthesis},
  author={Esser, Patrick and Kulal, Sumith and Blattmann, Andreas and Entezari, Rahim and M{\"u}ller, Jonas and Saini, Harry and Levi, Yam and Lorenz, Dominik and Sauer, Axel and Boesel, Frederic and Podell, Dustin and Dockhorn, Tim and English, Zion and Lacey, Kyle and Goodwin, Alex and Marek, Yannik and Rombach, Robin},
  booktitle={International Conference on Machine Learning (ICML)},
  year={2024}
}

@inproceedings{song2020denoising,
  title={Denoising diffusion implicit models},
  author={Song, Jiaming and Meng, Chenlin and Ermon, Stefano},
  booktitle={International Conference on Learning Representations},
  year={2020}
}

@article{guo2025deepseek,
  title={Deepseek-r1: Incentivizing reasoning capability in llms via reinforcement learning},
  author={Guo, Daya and Yang, Dejian and Zhang, Haowei and Song, Junxiao and Zhang, Ruoyu and Xu, Runxin and Zhu, Qihao and Ma, Shirong and Wang, Peiyi and Bi, Xiao and others},
  journal={arXiv preprint arXiv:2501.12948},
  year={2025}
}

@article{fu2025dynamic,
  title={Dynamic-TreeRPO: Breaking the Independent Trajectory Bottleneck with Structured Sampling},
  author={Fu, Xiaolong and Ma, Lichen and Guo, Zipeng and Zhou, Gaojing and Wang, Chongxiao and Dong, ShiPing and Zhou, Shizhe and Liu, Ximan and Fu, Jingling and Sin, Tan Lit and others},
  journal={arXiv preprint arXiv:2509.23352},
  year={2025}
}

@article{schulman2017proximal,
  title={Proximal policy optimization algorithms},
  author={Schulman, John and Wolski, Filip and Dhariwal, Prafulla and Radford, Alec and Klimov, Oleg},
  journal={arXiv preprint arXiv:1707.06347},
  year={2017}
}

@article{jiang2025t2i,
  title={T2i-r1: Reinforcing image generation with collaborative semantic-level and token-level cot},
  author={Jiang, Dongzhi and Guo, Ziyu and Zhang, Renrui and Zong, Zhuofan and Li, Hao and Zhuo, Le and Yan, Shilin and Heng, Pheng-Ann and Li, Hongsheng},
  journal={arXiv preprint arXiv:2505.00703},
  year={2025}
}

@article{kim2025tempflow,
  title={Tempflow-grpo: When timing matters for grpo in flow models},
  author={Kim, Jae Hyeon and Hwang, Sung Ju},
  journal={arXiv preprint arXiv:2509.16583},
  year={2025}
}

@article{zhang2025g2rpo,
  title={G2rpo: Granular grpo for precise reward in flow models},
  author={Zhang, Ziqiang and Wang, Hongli and Yang, Chenyang and Xia, Mengqi and Fang, Tao and Li, Bo and Zhu, Wenwu},
  journal={arXiv preprint arXiv:2510.11896},
  year={2025}
}

@article{black2023training,
  title={Training diffusion models with reinforcement learning},
  author={Black, Kevin and Janner, Michael and Du, Yilun and Kostrikov, Ilya and Levine, Sergey},
  journal={arXiv preprint arXiv:2305.13301},
  year={2023}
}

@article{ni2025treerpo,
  title={Treerpo: Tree relative policy optimization},
  author={Ni, Jiayi and Song, Ziqi and Liu, Jiaxin and Liu, Qian and Liu, Jian and Chen, Yimin and Chen, Ting and Huang, Gao},
  journal={arXiv preprint arXiv:2505.17208},
  year={2025}
}

@article{guo2025segment,
  title={Segment policy optimization: Effective segment-level credit assignment in rl for large language models},
  author={Guo, Yiran and Xu, Lijie and Liu, Jie and Ye, Dan and Qiu, Shuang},
  journal={arXiv preprint arXiv:2505.23564},
  year={2025}
}

@article{zhang2025tree,
  title={Tree-opo: Off-policy monte carlo tree-guided advantage optimization for multistep reasoning},
  author={Zhang, Hao and Liu, Zihan and Zhao, Yifan and Feng, Fangkai and Wang, William Yang and Pan, Sinong},
  journal={arXiv preprint arXiv:2505.21012},
  year={2025}
}

@inproceedings{Parrot,
author = {Lee, Seung Hyun and Li, Yinxiao and Ke, Junjie and Yoo, Innfarn and Zhang, Han and Yu, Jiahui and Wang, Qifei and Deng, Fei and Entis, Glenn and He, Junfeng and Li, Gang and Kim, Sangpil and Essa, Irfan and Yang, Feng},
title = {Parrot: Pareto-Optimal Multi-reward Reinforcement Learning Framework for Text-to-Image Generation},
year = {2024},
isbn = {978-3-031-72919-5},
publisher = {Springer-Verlag},
address = {Berlin, Heidelberg},
url = {https://doi.org/10.1007/978-3-031-72920-1_26},
doi = {10.1007/978-3-031-72920-1_26},
booktitle = {Computer Vision – ECCV 2024: 18th European Conference, Milan, Italy, September 29–October 4, 2024, Proceedings, Part XXXVIII},
pages = {462–478},
numpages = {17},
location = {Milan, Italy}
}

@inproceedings{huang2023t2i,
  title={T2i-compbench: A comprehensive benchmark for open-world compositional text-to-image generation},
  author={Huang, Kaiyi and Sun, Kaiyue and Xie, Enze and Li, Zhenguo and Liu, Xihui},
  booktitle={Advances in Neural Information Processing Systems},
  volume={36},
  year={2023}
}

@inproceedings{chen2023textdiffuser,
  title={Textdiffuser: Diffusion models as text painters},
  author={Chen, Jingye and Huang, Yupan and Lv, Tengchao and Cui, Lei and Chen, Qifeng and Wei, Furu},
  booktitle={Advances in Neural Information Processing Systems},
  volume={36},
  year={2023}
}

@article{yang2025qwen3,
  title={Qwen3 technical report},
  author={Yang, An and Li, Anfeng and Yang, Baosong and Zhang, Beichen and Hui, Binyuan and Zheng, Bo and Yu, Bowen and Gao, Chang and Huang, Chengen and Lv, Chenxu and others},
  journal={arXiv preprint arXiv:2505.09388},
  year={2025}
}

@misc{chen2025r1v,
  author       = {Chen, Liang and Li, Lei and Zhao, Haozhe and Song, Yifan and Vinci},
  title        = {R1-V: Reinforcing Super Generalization Ability in Vision-Language Models with Less Than \$3},
  howpublished = {\url{https://github.com/Deep-Agent/R1-V}},
  note         = {Accessed: 2025-02-02},
  year         = {2025}
}

@article{kirstain2023pick,
  title={Pick-a-pic: An open dataset of user preferences for text-to-image generation},
  author={Kirstain, Yuval and Polyak, Adam and Singer, Uriel and Matiana, Shahbuland and Penna, Joe and Levy, Omer},
  journal={Advances in neural information processing systems},
  volume={36},
  pages={36652--36663},
  year={2023}
}

@misc{wang2022gitgenerativeimagetotexttransformer,
      title={GIT: A Generative Image-to-text Transformer for Vision and Language}, 
      author={Jianfeng Wang and Zhengyuan Yang and Xiaowei Hu and Linjie Li and Kevin Lin and Zhe Gan and Zicheng Liu and Ce Liu and Lijuan Wang},
      year={2022},
      eprint={2205.14100},
      archivePrefix={arXiv},
      primaryClass={cs.CV},
      url={https://arxiv.org/abs/2205.14100}, 
}
}

\clearpage
\setcounter{section}{0}
\renewcommand\thesection{\Alph{section}}
\renewcommand\theHsection{\Alph{section}} 

\section{Unscaled Reward-Based Grouping}
\label{sec:supp-unscaled-reward}

In the main paper (Section~3.3), we introduce \emph{reward-based grouping}, where advantages are first computed independently for each reward type and then aggregated:
\begin{equation}
    \hat{A}^i_m = \frac{R^i_m - \mu_m}{\sigma_m},
\end{equation}
\begin{equation}
    \hat{A}^i = \frac{1}{M}\sum_{m=1}^M w_m \hat{A}^i_m.
\end{equation}
The $1/M$ factor is a natural scaling choice that keeps the overall gradient magnitude roughly stable as the number of reward terms $M$ increases, helping to avoid overly large updates when $M$ varies across tasks.

From an optimization standpoint, however, this global scaling is not always required. In particular, when $M$ is small and fixed (as in our experiments with three rewards), the overall gradient scale can often be absorbed into the learning-rate schedule. To verify this, we also experiment with an \emph{unscaled} variant that removes the $1/M$ factor and directly sums the weighted per-reward advantages:
\begin{equation}
    \hat{A}^i_{\text{unscaled}} = \sum_{m=1}^M w_m \hat{A}^i_m.
\end{equation}

Table~\ref{tab:supp_unscaled_reward} reports results on the OCR-Color-10 multi-objective setting with color, OCR, and PickScore rewards. We compare (i) the scaled reward-based grouping used in the main paper as our default \emph{Multi-GRPO} configuration and (ii) the unscaled variant that does not divide by $M$. For reference, we also include the original \texttt{Flow-GRPO} baseline. On this dataset, the unscaled variant achieves higher performance across all reward dimensions. Nonetheless, for better scalability to larger $M$ and to keep the aggregated advantage well-scaled as we add more reward terms, we adopt the scaled version in the main paper and include the unscaled variant here for completeness.

\paragraph{Intuition.}
When advantage signals from different rewards conflict, the aggregated gradient can be noisy; in this case, multiplying by $1/M$ mainly acts as a global damping factor that reduces the influence of this noisy guidance without changing its direction. When advantage signals from different rewards agree, removing $1/M$ preserves a larger gap between high- and low-advantage samples and can strengthen the alignment signal. In our three-reward setting, this stronger signal appears to be helpful, but this may be specific to the dataset and training setup we study here. We believe that more systematic experiments across different models, datasets, and numbers of rewards would be valuable for understanding when including the $1/M$ factor is beneficial.

\begin{table}[h]
\centering
\caption{
Reward-based grouping with and without the $1/M$ scaling factor on the \textit{OCR-Color-10} multi-reward setting. 
Flow-GRPO is the original baseline without reward-based grouping; ``Multi-GRPO (main paper)'' uses the $1/M$ scaling, while ``Multi-GRPO (unscaled)'' removes it under otherwise identical settings.
}
\label{tab:supp_unscaled_reward}
\begin{tabular}{lccc}
\toprule
Method & $R_{\text{color}}$ $\uparrow$ & $R_{\text{ocr}}$ $\uparrow$ & $R_{\text{pick}}$ $\uparrow$ \\
\midrule
Flow-GRPO  & 0.9471 & 0.9428 & 0.8689 \\
Multi-GRPO (main paper)  & 0.9516 & 0.9382 & 0.8966 \\
Multi-GRPO (unscaled)  &  \textbf{0.9840} & \textbf{0.9549} & \textbf{0.9147} \\
\bottomrule
\end{tabular}
\end{table}

\section{Branching Rules from State and Action Perspectives: Timesteps vs.\ Denoising Steps}
\label{sec:supp-branch-impl}

In Sec.~3.2 of the main paper, we define the branching rule in terms of
\emph{diffusion timesteps} (state indices)
\[
\mathcal{B} = \{b_1{:}B_1, b_2{:}B_2, \ldots, b_K{:}B_K\},
\]
where each $b_k \in \{T,T{-}1,\ldots,0\}$ indexes a state $\mathbf{x}_{b_k}$ and
$B_k$ is the number of children each active trajectory branches into at that
timestep. 

From an implementation perspective, it is more natural to think in terms of
\emph{denoising steps} (actions): step $i = 1,\ldots,T$ denotes the transition
from $\mathbf{x}_{T-i+1}$ to $\mathbf{x}_{T-i}$. In code we simply write
“step $i$” for this update. When describing implementation details in the
paper, we instead refer to the same denoising step $i$ by its \emph{target
timestep} $t=T-i$, e.g., “the denoising step $T{-}2$” means the update
$\mathbf{x}_{T-1} \rightarrow \mathbf{x}_{T-2}$. This keeps the notation
consistent with the reverse-time indexing $T, T{-}1, \ldots, 0$ used in the
method section.

For example, the schedule used in the implementation detail
$\mathcal{B} = \{T{-}2{:}3,\ T{-}4{:}2\}$
means we branch when producing $\mathbf{x}_{T-2}$ and $\mathbf{x}_{T-4}$;
these are the 2nd and 4th denoising steps from the end, so the code-level rule
is \texttt{branch\_rule = \{2:3, 4:2\}}.

In short, timesteps $b_k$ index \emph{states} and are used in the method
description, while denoising steps $i$ index \emph{actions} and are used in the
implementation; the two notations are just different views of the same
branching tree.

\begin{figure*}[t]
    \centering
    \includegraphics[width=\linewidth]{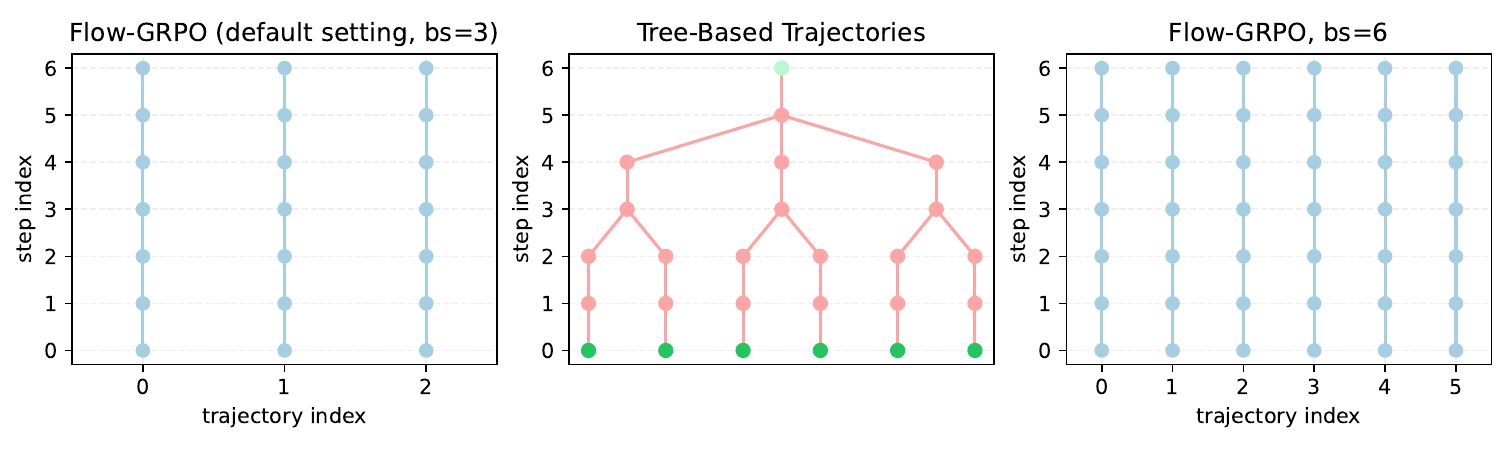}
    \caption{Denoising trajectories in the FLUX single-reward experiment. From left to right: default \texttt{Flow-GRPO} with $\text{bs}=3$ (three independent trajectories), our tree-based trajectories using branching rule $\{T{-}2{:}3,\ T{-}4{:}2\}$, and enlarged-batch \texttt{Flow-GRPO} with $\text{bs}=6$ (six independent trajectories).}
    \label{fig:tree_sup}
\end{figure*}

\section{Early vs.\ Late Branching in Tree-Based Trajectories Without Reward-Based Grouping}

In the main paper we study the effect of the branching rule in the \emph{full}
Multi-GRPO setting, where tree-based trajectories are combined with
reward-based grouping. That ablation compares an ``early'' branching rule
$\{T{-}2{:}3,\ T{-}4{:}2\}$ with a ``late'' branching rule
$\{T{-}3{:}2,\ T{-}7{:}3\}$, and finds that the former performs better across
all three rewards.

Here we isolate the role of the branching rule itself by turning off
reward-based grouping and keeping only the tree-structured rollouts. On the OCR-Color-10 task, we fix all training hyperparameters and reward
weights, train every model for $4$k steps, and vary only the branching rule.
Specifically, we compare our default early-branching rule $\{T{-}2{:}3,\ T{-}4{:}2\}$ with a
slightly later-branching variant that moves the second branching point from $T{-}4$
to $T{-}7$, i.e., uses the rule $\{T{-}2{:}3,\ T{-}7{:}2\}$, while keeping the total
number of leaf trajectories unchanged.

As shown in Table~\ref{tab:supp_branch_sd3}, the default rule again yields
higher color, OCR, and PickScore rewards than the alternative. This indicates
that the benefit of branching at earlier (higher-noise) steps comes from the
tree-based trajectories itself, and is largely independent of whether
reward-based grouping is applied.

\begin{table}[h]
\centering
\caption{Comparison of branching rules on the OCR-Color-10 setting using \emph{tree-based trajectories only} (no reward-based grouping). All models are trained for 4k steps with the same hyperparameters; only the branching rule differs.}
\label{tab:supp_branch_sd3}
\begin{tabular}{lccc}
\toprule
Branching Rule & $R_{\text{color}}$ $\uparrow$ & $R_{\text{ocr}}$ $\uparrow$ & $R_{\text{pick}}$ $\uparrow$ \\
\midrule
$\{T{-}2{:}3,\ T{-}7{:}2\}$  & 0.9502 & 0.9338 & 0.8746 \\
$\{T{-}2{:}3,\ T{-}4{:}2\}$  & \textbf{0.9612} & \textbf{0.9462} & \textbf{0.8890} \\
\bottomrule
\end{tabular}
\end{table}

\section{Tree-Based Trajectories vs.\ Flow-GRPO with Larger Batch Size}
\label{sec:supp-fair-bs}

One may wonder whether the gains of our tree-based trajectories are simply due to performing more `denoising steps'' per prompt.
To address this concern, we run a controlled single-reward experiment on the FLUX.1-Dev model using PickScore only.

The default \texttt{Flow-GRPO} baseline uses batch size $\text{bs}=3$ on $8$ GPUs, giving $8 \times 3 = 24$ trajectories per batch. In \texttt{Flow-GRPO} baseline, each trajectory follows a independent denoising path from step $T$ down to $0$. We compare three settings: the original $\text{bs}=3$ configuration, an enlarged-batch baseline with $\text{bs}=6$, and our tree-based trajectories with $\text{bs}=1$.

Our method (tree-based trajectories), illustrated in the right panel of Fig.~\ref{fig:tree_sup}, branches at timestep 5 and timestep 3, resulting in
$1 \times 2 \times 3 = 6$ leaf nodes per prompt. The total numbers of denoising
steps per gpu are as follows: (1) \texttt{Flow-GRPO} with $\text{bs}=3$ uses
$18$ steps ($3\times6 = 18$),
(2) \texttt{Flow-GRPO} with $\text{bs}=6$ uses $36$ steps ($6\times6=36$), (3) and our method uses $25$ steps ($1 + 2\times3 + 6\times3 = 25$). Thus, the $\text{bs}=6$ variant performs the most
denoising updates, our method uses fewer updates while still producing
six leaves, and the $\text{bs}=3$ baseline uses the fewest.

Figure~\ref{fig:tree_sup} visualizes the corresponding denoising trajectories. Table~\ref{tab:supp_flux_pick} reports the quantitative results: even after increasing the batch size to $\text{bs}=6$, the tree-structured sampling achieves a higher PickScore reward while using fewer denoising steps than the enlarged-batch \texttt{Flow-GRPO} baseline, suggesting that tree-structured rollout is beneficial for improving performance beyond simply performing more denoising steps.

\begin{table}[h]
\centering
\caption{FLUX single-reward experiment (PickScore only): enlarged-batch \texttt{Flow-GRPO} vs.\ our tree-based trajectories. Our method attains the highest PickScore while using fewer denoising steps than the $\text{bs}=6$ baseline.}
\label{tab:supp_flux_pick}
\begin{tabular}{l c}
\toprule
Method & PickScore $\uparrow$ \\
\midrule
\texttt{Flow-GRPO}, bs = 3   &  23.65\\
\texttt{Flow-GRPO}, bs = 6   & 24.02 \\
Tree schedule ($\{T{-}2{:}3,\ T{-}4{:}2\}$) &  \textbf{24.64}\\
\bottomrule
\end{tabular}
\end{table}

\begin{figure*}[ht]
    \centering
    \includegraphics[width=\linewidth]{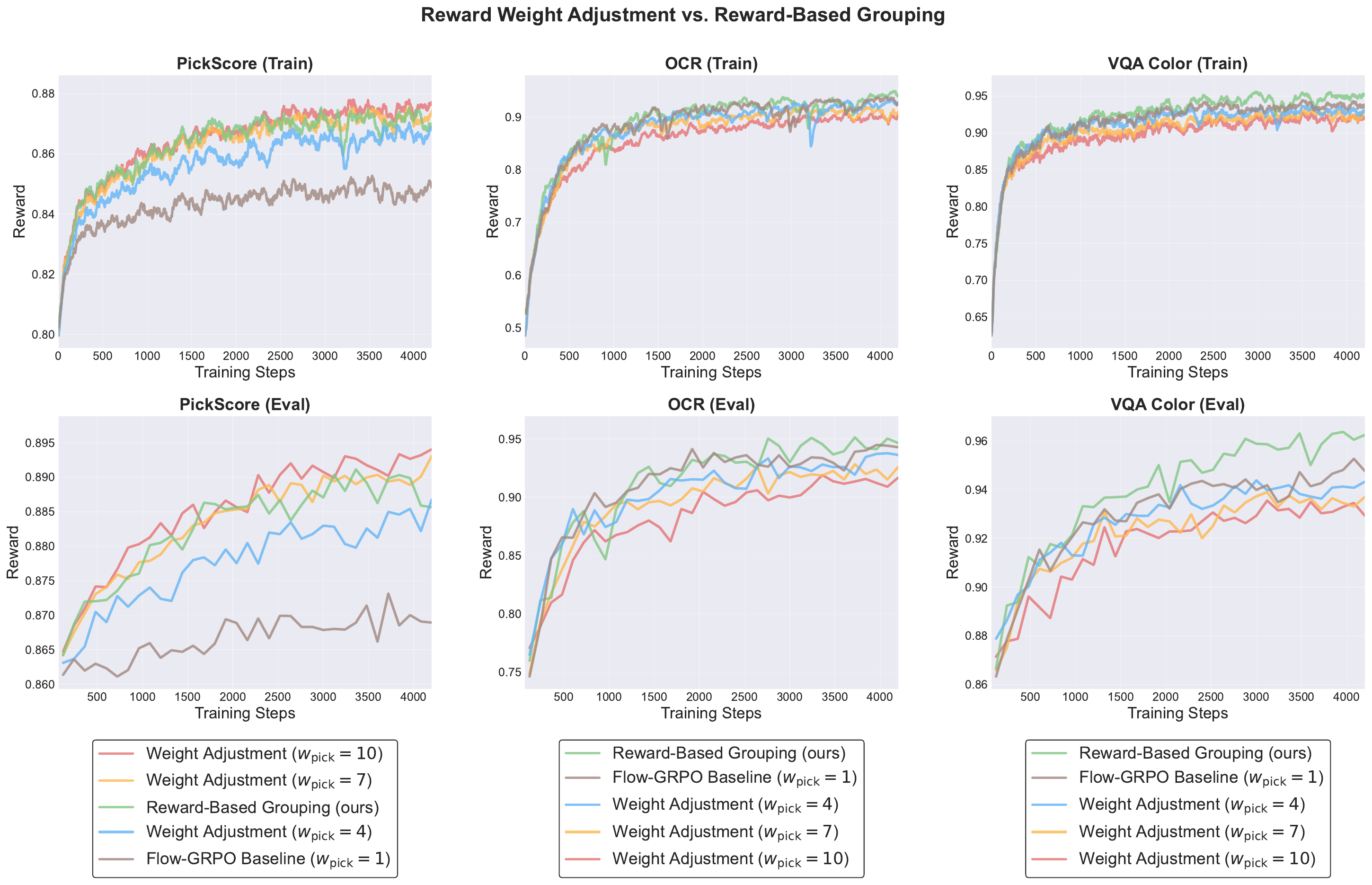}
    \caption{
        \textbf{Reward weight adjustment vs.\ reward-based grouping on \textit{OCR-Color-10}.}
        We compare \texttt{Flow-GRPO} baseline
        ($w_{\text{ocr}}{=}w_{\text{color}}{=}w_{\text{pick}}{=}1$),
        three \emph{reward weight adjustment} variants that increase only the
        PickScore weight ($w_{\text{pick}} \in \{4,7,10\}$ with
        $w_{\text{ocr}}{=}w_{\text{color}}{=}1$), and our
        \emph{reward-based grouping} method (all weights fixed to $1$ with
        per-reward normalization).
        Simply increasing $w_{\text{pick}}$ raises $R_{\text{pick}}$ but consistently
        harms $R_{\text{ocr}}$ and $R_{\text{color}}$, whereas reward-based
        grouping achieves the best $R_{\text{ocr}}$ and $R_{\text{color}}$ while
        maintaining competitive $R_{\text{pick}}$.
    }
    \label{fig:pick_weight_ablation}
\end{figure*}
\section{Reward Weight Adjustment vs.\ Reward-Based Grouping}
\label{sec:supp-weight-ablation}

This section asks whether the effect of reward-based grouping can be replicated
by simply retuning the scalar reward weights in a standard weighted sum.
On the three-reward \textit{OCR-Color-10} setting we use
\begin{equation}
R_{\text{total}}
= w_{\text{ocr}} R_{\text{ocr}}
+ w_{\text{color}} R_{\text{color}}
+ w_{\text{pick}} R_{\text{pick}},
\label{eq:total_reward}
\end{equation}

The original \texttt{Flow-GRPO} uses equal weights
$w_{\text{ocr}} = w_{\text{color}} = w_{\text{pick}} = 1$.
We treat this as the baseline and then construct three
\emph{reward weight adjustment} variants by keeping
$w_{\text{ocr}} = w_{\text{color}} = 1$ and changing only the PickScore weight
to $w_{\text{pick}} \in \{4,7,10\}$. Our reward-based grouping method uses
$w_{\text{ocr}} = w_{\text{color}} = w_{\text{pick}} = 1$, but compute advantages
via per-reward normalization.

The behaviour of reward weight adjustment is intuitive: as $w_{\text{pick}}$
increases, the PickScore reward $R_{\text{pick}}$ improves, but this consistently
comes at the expense of $R_{\text{ocr}}$ and $R_{\text{color}}$, which both
decrease for larger $w_{\text{pick}}$. In contrast, reward-based grouping
achieves a more balanced outcome: it attains higher $R_{\text{pick}}$ than the
low-weighted variant ($w_{\text{pick}}{=}4$) while remaining competitive with more
aggressively up-weighted runs, and at the same time yields the best
$R_{\text{ocr}}$ and $R_{\text{color}}$ among all variants. This indicates that
normalizing each reward dimension separately is not equivalent to adjusting a
single scalar weight, and actually gives a better multi-objective trade-off.

Figure~\ref{fig:pick_weight_ablation} visualizes this comparison by plotting
training rewards and evaluation rewards for the
\texttt{Flow-GRPO} baseline ($w_{\text{pick}}{=}1$), the three reward weight
adjustment variants ($w_{\text{pick}} \in \{4,7,10\}$), and our reward-based
grouping method.

\section{Qualitative Comparison on the \textit{OCR-Color-10} Dataset}
\label{sec:supp-ocr-qual}

We first present qualitative results on the multi-objective \textit{OCR-Color-10} dataset (Task~2), where SD3.5-M is optimized with three rewards: text fidelity $R_{\text{ocr}}$, text--color consistency $R_{\text{color}}$, and image quality $R_{\text{pick}}$. For each prompt, we compare the original SD3.5-M, the \texttt{Flow-GRPO} baseline, and our method.

Across diverse prompts, the base model SD3.5-M already produces globally plausible scenes (lighting, composition, and color palettes), but often fails on the text: \textbf{the rendered text may be incomplete or distorted}, contain spelling errors, or appear in a color that does not match the instruction. 

The \texttt{Flow-GRPO} baseline substantially improves legibility and correctness: the rendered strings are more often readable and match the target phrase. However, the text layouts tend to become very regular and “boxy” (e.g., flat, centered overlays), which \textbf{often look pasted onto the image instead of blending naturally with the background geometry and perspective, thereby reducing perceived realism.}

\begin{figure*}[t]
    \centering
    \includegraphics[
        width=\linewidth,
        height=0.96\textheight,
        keepaspectratio
    ]{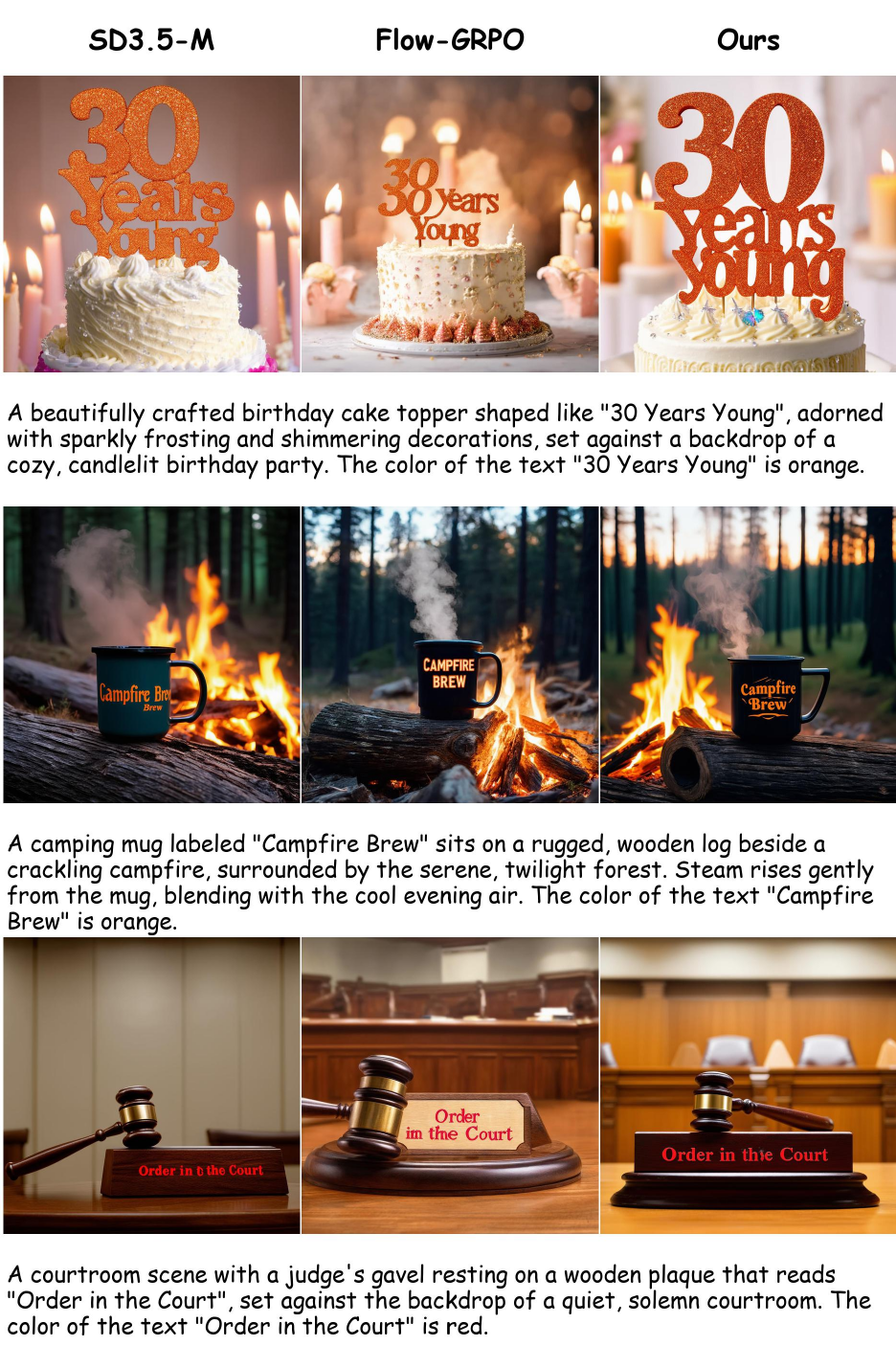}
    \caption{Qualitative comparison on the \textit{OCR-Color-10} dataset (part~I). 
    For each prompt, we show SD3.5-M (left), \texttt{Flow-GRPO} (middle), and our method (right). 
    Our approach produces text that is both legible and well integrated into the scene, 
    while more faithfully following the textual and color instructions and yielding images 
    with higher perceived realism and aesthetics.}
    \label{fig:ocr_color_qual_a}
\end{figure*}

\begin{figure*}[t]
    \centering
    \includegraphics[
        width=\linewidth,
        height=0.96\textheight,
        keepaspectratio
    ]{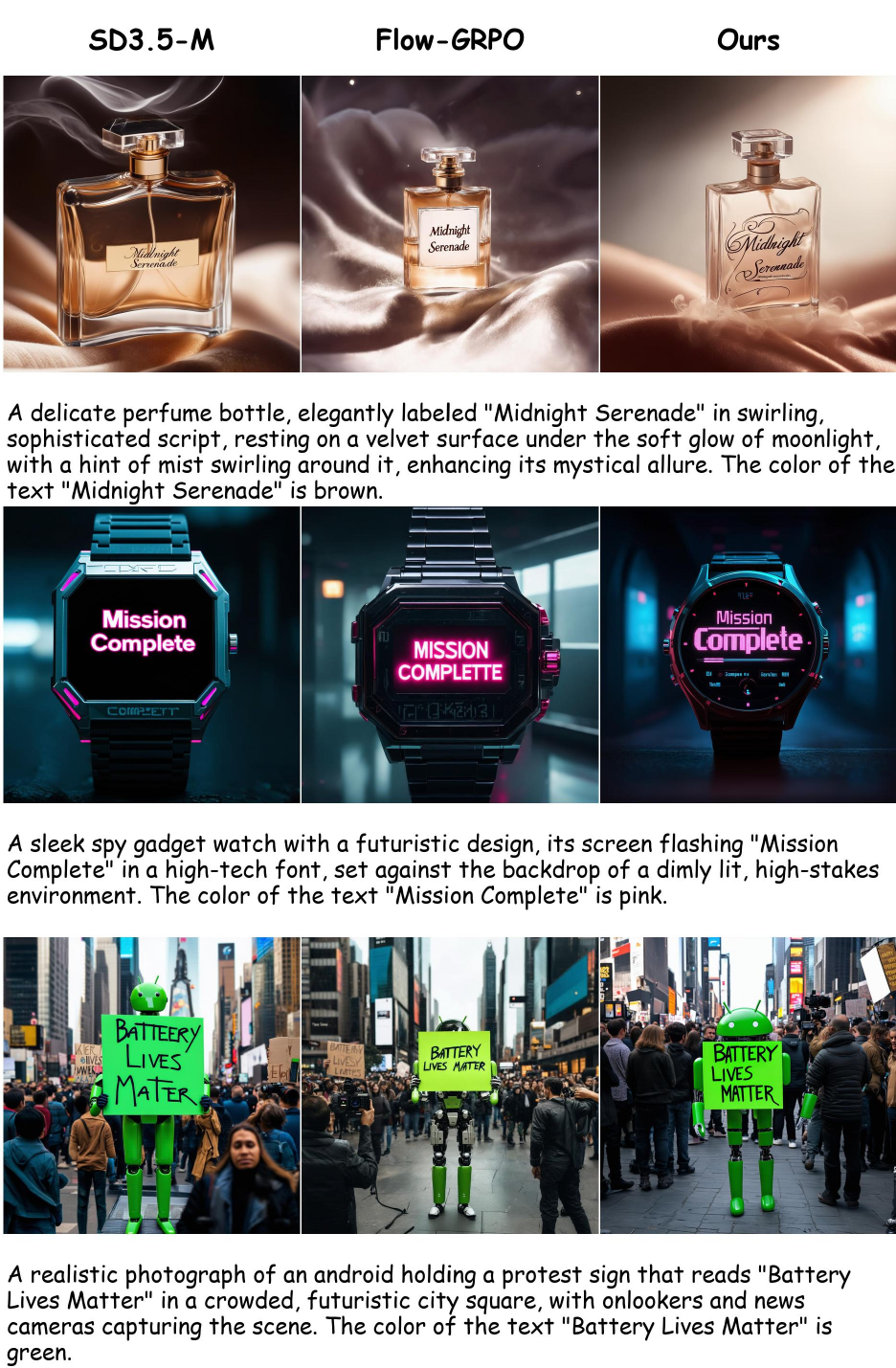}
    \caption{Qualitative comparison on the \textit{OCR-Color-10} dataset (part~II). 
    Additional examples under the same setting as Figure~\ref{fig:ocr_color_qual_a}.}
    \label{fig:ocr_color_qual_b}
\end{figure*}

\begin{figure*}[t]
    \centering
    \includegraphics[
        width=\linewidth,
        height=0.96\textheight,
        keepaspectratio
    ]{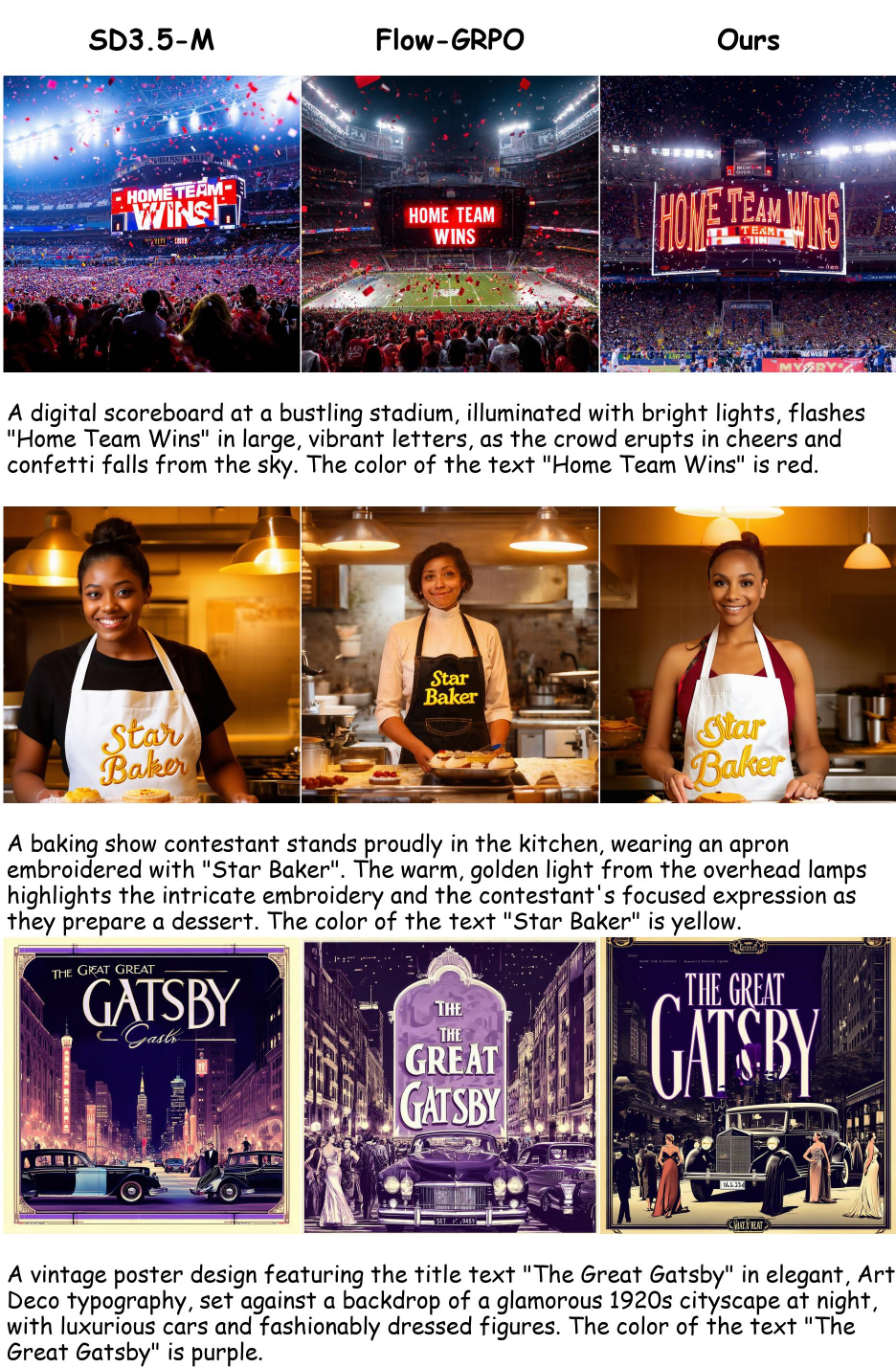}
    \caption{Qualitative comparison on the \textit{OCR-Color-10} dataset (part~III). 
    Additional examples under the same setting as Figure~\ref{fig:ocr_color_qual_a}.}
    \label{fig:ocr_color_qual_c}
\end{figure*}

\begin{figure*}[t]
    \centering
    \includegraphics[
        width=\linewidth,
        height=0.98\textheight,
        keepaspectratio
    ]{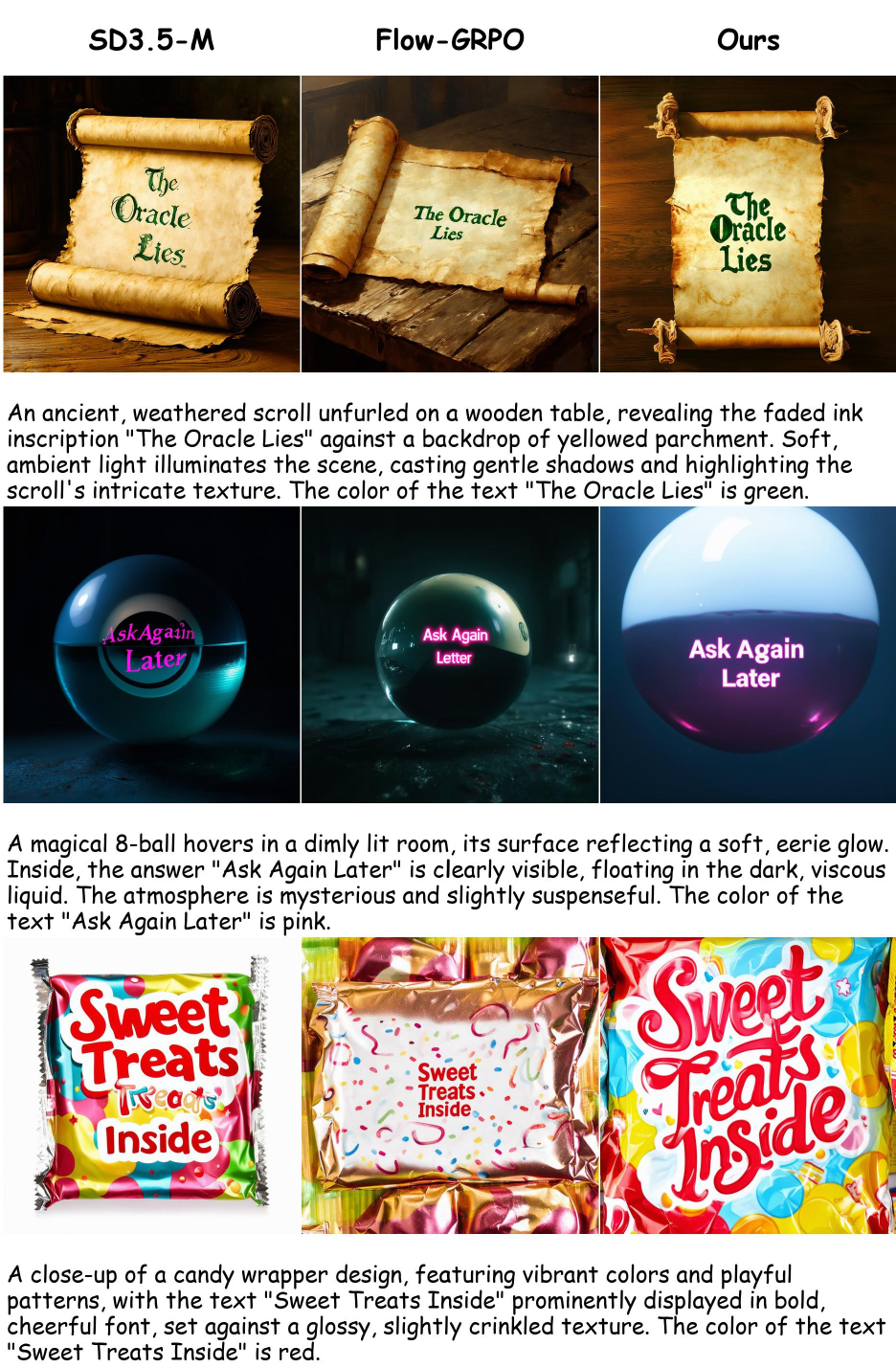}
    \caption{Qualitative comparison on the \textit{OCR-Color-10} dataset (part~IV). 
    Additional examples under the same setting as Figure~\ref{fig:ocr_color_qual_a}.}
    \label{fig:ocr_color_qual_d}
\end{figure*}

Our method combines the strengths of both. As illustrated in Figures~\ref{fig:ocr_color_qual_a}--\ref{fig:ocr_color_qual_d}, the generated text is typically correct and clearly legible, while also intergrating more naturally with the background in terms of viewpoint and lighting. The text color more reliably matches the specified color, and \textbf{the words appear visually embedded in the scene rather than simply overlaid on top of the image}. These qualitative trends are consistent with the quantitative gains reported in the main paper.

\section{Qualitative Comparison on the \textit{PickScore-25k} Dataset}
\label{sec:supp-pickscore-qual}

We next show qualitative results on the single-objective \textit{PickScore-25k} dataset (Task~1), where FLUX.1-Dev is aligned using a single aesthetic reward $R_{\text{pick}}$. We compare three models: the original FLUX.1-Dev, the \texttt{Flow-GRPO} baseline, and our method with tree-based trajectories (no reward-based grouping is needed since there is only one reward). To mitigate reward-hacking effects that can arise in single-reward settings, all qualitative results are taken from an early training checkpoint (step 1020), where the images still maintain reasonable visual quality and no noticeable degradation in image quality is observed.

The base FLUX.1-Dev model already generates visually pleasing images, but it \textbf{tends to produce images with less fine-grained detail and weaker overall composition}. Some samples look slightly less refined or lack a clear focus. The \texttt{Flow-GRPO} baseline improves PickScore and generally produces sharper textures and cleaner object boundaries, but \textbf{can still miss some aspects of the prompt}.

\begin{figure*}[t]
    \centering
    \includegraphics[
        width=\linewidth,
        height=0.96\textheight,
        keepaspectratio
    ]{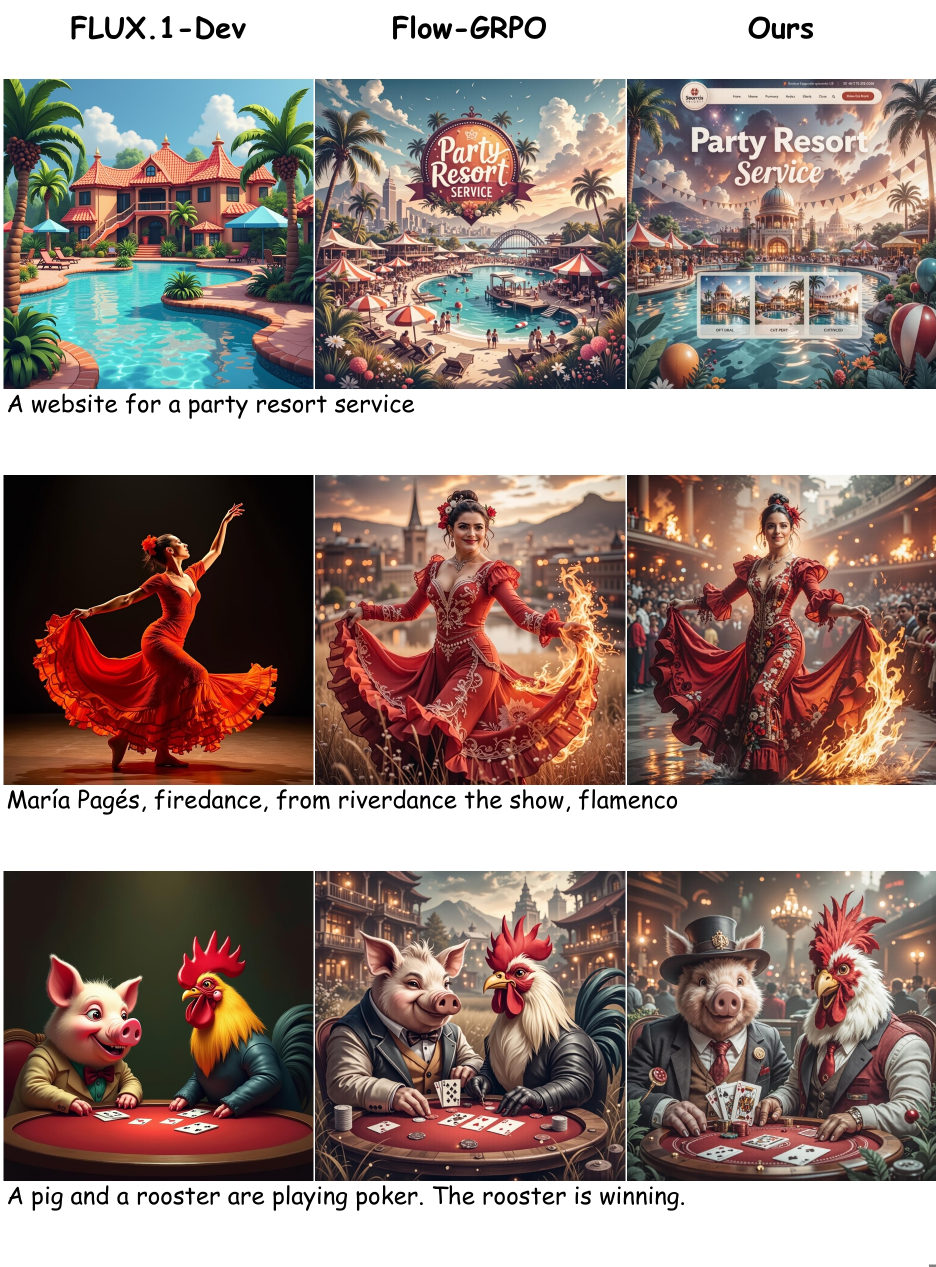}
    \caption{Qualitative comparison on the \textit{PickScore-25k} dataset (part~I). For each prompt, we show FLUX.1-Dev (left), \texttt{Flow-GRPO} (middle), and our method (right). Our approach yields images with higher human-perceived aesthetics, featuring richer local details. Moreover, it demonstrates stronger adherence to text prompts, consistent with the quantitative PickScore results.}
    \label{fig:pickscore_flux_qual_a}
\end{figure*}

\begin{figure*}[t]
    \centering
    \includegraphics[
        width=\linewidth,
        height=0.96\textheight,
        keepaspectratio
    ]{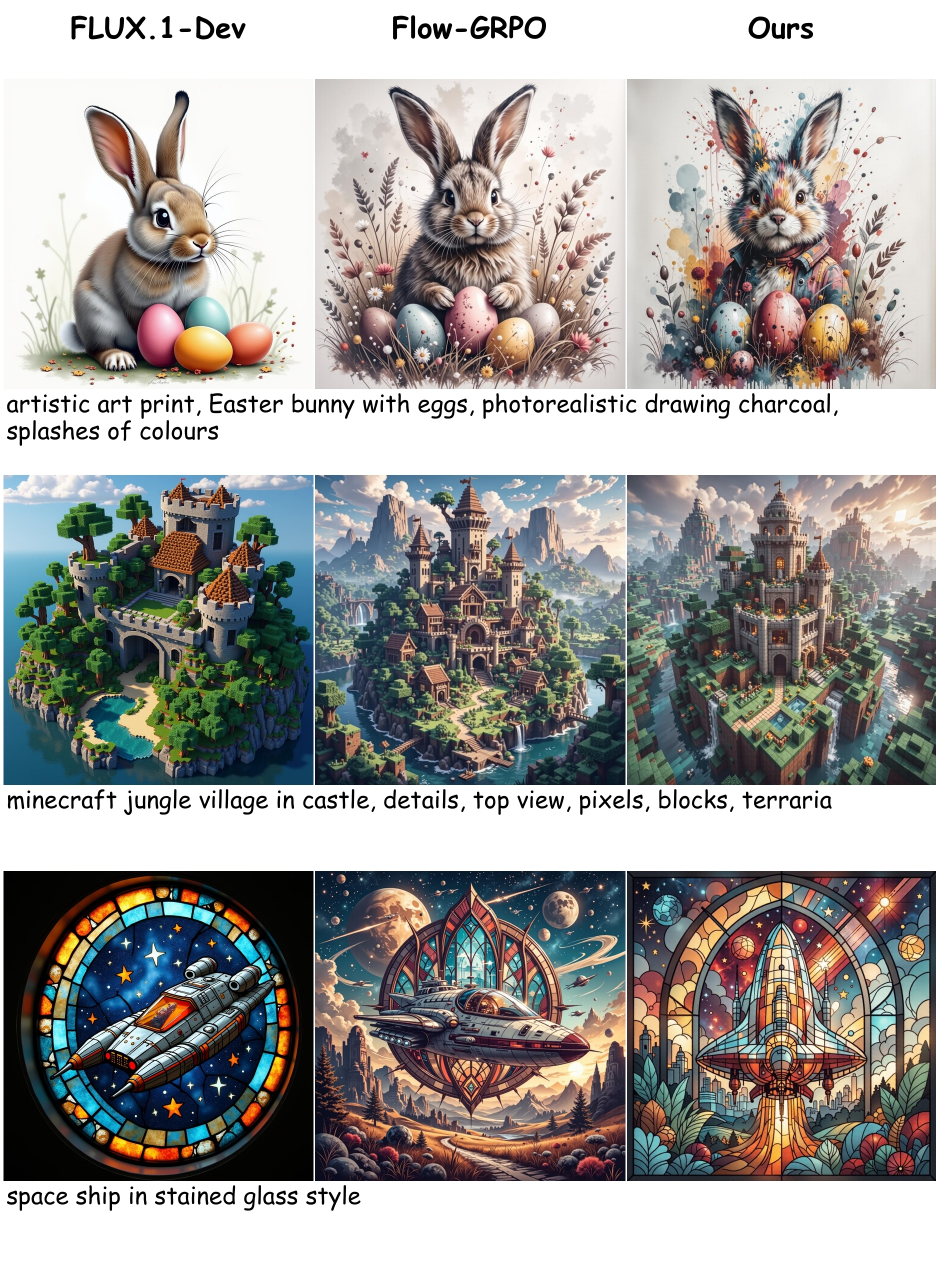}
    \caption{Qualitative comparison on the \textit{PickScore-25k} dataset (part~II). The same ordering is used as in Figure~\ref{fig:pickscore_flux_qual_a}: FLUX.1-Dev (left), \texttt{Flow-GRPO} (middle), and our method (right). }
    \label{fig:pickscore_flux_qual_b}
\end{figure*}

Our method further improves both visual quality and prompt adherence. In Figures~\ref{fig:pickscore_flux_qual_a} and~\ref{fig:pickscore_flux_qual_b}, our samples typically exhibit richer fine-grained structures and better visual quality, while avoiding an overprocessed appearance. The generated images also more faithfully reflect the intended content. For instance, in the first row of Figure~\ref{fig:pickscore_flux_qual_a}, corresponding to the prompt ``A website for a party resort service'', \textbf{neither the base FLUX.1-Dev model nor the Flow-GRPO baseline produces an image that clearly reads as a website, whereas our method explicitly renders a website-style layout}. Overall, these qualitative comparisons align with the quantitative results in the main paper, where our method achieves the highest PickScore.

\section{Limitations and Future Work}
Unlike recent approaches~\cite{li2025mixgrpo,zhang2025g2rpo} that pursue efficiency by hybridizing SDE–ODE sampling and training on a few key denoising steps, our motivation for introducing a tree-based trajectory by branching at early steps primarily lies in accurately estimate the potential of early denoising steps. Thus, we still follow Flow-GRPO’s~\cite{liu2025flowgrpo} SDE sampling strategy to ensure that later denosing steps can be effectively explored during training. In practice, our framework is also compatible with hybrid SDE–ODE designs—for example, using SDE sampling near branching nodes $b_k$ (i.e., from $b_k+1$ to $b_k$, and from $b_k$ to $b_k-1$) and ODE sampling elsewhere to further accelerate training. In this case, to encourage broader exploration across timesteps, branching steps could be randomly selected from early steps. 
We leave such integration and exploration for future work.

\end{document}